\documentclass[preprint,review,12pt]{elsarticle}

\usepackage[labelfont=bf,singlelinecheck=false]{caption}
\usepackage{amssymb}

\usepackage{lineno}

\usepackage{hyperref}
\hypersetup{
    colorlinks=True,
    pdfborder={0 0 0},
}
\usepackage{multirow}
\usepackage{graphicx}
\usepackage{adjustbox}
\usepackage{tabularx,booktabs}
\newcolumntype{C}{>{\centering\arraybackslash}X} 
\usepackage{lipsum}
\usepackage{algorithm}
\usepackage{algpseudocode}
\usepackage{hyperref}
\usepackage{url}
\usepackage{amsmath} 
\usepackage[dvipsnames]{xcolor}
\usepackage{array}
\usepackage[font=small,skip=4pt]{caption}

\journal{Computers and Electronics in Agriculture}

\begin{document}


\begin{frontmatter}



\title{Cloud gap-filling with deep learning for improved grassland monitoring}
\author[1]{Iason Tsardanidis\corref{cor1}}\ead{j.tsardanidis@noa.gr}
\author[2]{Alkiviadis Koukos}
\author[3]{Vasileios Sitokonstantinou\corref{cor1}}\ead{vasileios.sitokonstantinou@uv.es}
\author[1]{Thanassis Drivas}
\author[1]{Charalampos Kontoes}

\cortext[cor1]{Corresponding authors.}



\affiliation[1]{organization={Operational Unit BEYOND Centre for EO Research and Satellite Remote Sensing, Institute for Astronomy, Astrophysics Space Applications and Remote Sensing (IAASARS), National Observatory of Athens},
            postcode={11523},
            state={Athens},
            country={Greece}}  
\affiliation[2]{organization={DHI Water \& Environment},
            postcode={2970}, 
            state={Hørsholm},
            country={Denmark}}         
\affiliation[3]{organization={Image Processing Laboratory (IPL), Parc Científic, Universitat de València},
            postcode={46980 Paterna}, 
            state={València},
            country={Spain}}

\begin{abstract}

Uninterrupted optical image time series are crucial for the timely monitoring of agricultural land changes, particularly in grasslands. However, the continuity of such time series is often disrupted by clouds. In response to this challenge, we propose an innovative deep learning method that integrates cloud-free optical (Sentinel-2) observations and weather-independent (Sentinel-1) Synthetic Aperture Radar (SAR) data. 
Our approach employs a hybrid architecture combining Convolutional Neural Networks (CNNs) and Recurrent Neural Networks (RNNs) to generate continuous Normalized Difference Vegetation Index (NDVI) time series, highlighting the role of NDVI in the synergy between SAR and optical data. We demonstrate the significance of observation continuity by assessing the impact of the generated NDVI time series on the downstream task of grassland mowing event detection. We conducted our study in Lithuania, a country characterized by extensive cloud coverage, and compared our approach with alternative interpolation techniques (i.e., linear, Akima, quadratic). Our method outperformed these techniques, achieving an average Mean Absolute Error (MAE) of 0.024 and a coefficient of determination ($\mathrm{R^2}$) of 0.92. 
Additionally, our analysis revealed improvement in the performance of the mowing event detection, with F1-score up to 84\% using two widely applied mowing detection methodologies. 
Our method also effectively mitigated sudden shifts and noise originating from cloudy observations, which are often missed by conventional cloud masks and adversely affect mowing detection precision.

\end{abstract}






\end{frontmatter}

\section{Introduction}\label{sec:introduction}

Grasslands cover around one-third of the global agricultural land and they play a key role in food production, animal biodiversity, and global carbon cycle \citep{mara2012}. They provide a wide range of ecosystem services, such as (i) provision of fodder for livestock, (ii) wildlife habitats, (iii) filtering or retention functions of waterways and greenhouse gas emissions, (iv) carbon storage, (v) pest control, (vi) crop pollination and (vii) protection against soil erosion \citep{zhao2020grassland}. The 2013 reform of the Common Agricultural Policy (CAP) has emphasized the importance of grassland use intensity by requiring a minimum number of mowing or grazing events. Therefore, national paying agencies have been tasked with monitoring compliance with this minimum activity condition. This information is important for the consolidation of agro-ecological measures and conservation schemes with regards to climate change \citep{d'Adrimont2018,reinermann2023}. 

The ever-increasing availability of remote sensing data has enabled the development of numerous competent agricultural monitoring systems. Most studies on grassland monitoring use remote sensing to address the problem of discriminating the different grassland categories (based on land cover, use intensity, etc.) and evaluate the various management strategies \citep{ali2016,Reinermann2020}. However, few focus on the quantification of grassland grazing intensity and the detection of mowing events. In most cases, optical Earth observation data (MODIS, RapidEye, Landsat 8, Sentinel-2, etc.) are exploited \citep{gomez2017,estel2018}. Several studies evaluate the mapping of the mowing or grazing frequency using statistical methods over multi-spectral image time series on a regional or national scale \citep{kolecka2018,Griffiths2019,Stumpf2020}. Therefore, relevant vegetation indices, such as the Normalized Difference Vegetation Index (NDVI), the Leaf Area Index (LAI), or others, are used to detect drastic changes in the grassland vegetation cover, between consecutive observations, which is related to the amount of biomass removed.

Nevertheless, one of the main limitations of optical data is missing or invalid values caused by cloud coverage and unfavorable weather conditions. This problem is more evident in northern countries, where even the short revisit time of 5 days of Sentinel-2 can result in time series with long gaps \citep{sudmanns2019}.
Consequently, frequent cloud cover in these areas obstructs the detection of mowing practices, rendering algorithms reliant on optical data less effective in discerning these events \citep{kolecka2018}. 
Hence, there is a need to artificially reconstruct these missing or corrupted values. This reconstruction can be performed by using only optical data, only Synthetic Aperture Radar (SAR) data, or collaborative exploitation of both \citep{li2022}. 

The most straightforward method for addressing gaps in optical data is to use temporal gap-filling techniques that rely only on optical data, such as the linear (or splines) interpolation \citep{zeng2013}. Even though these methods have proved sufficient for several tasks \citep{fauvel2019,sitok2021}, in cases where we need to precisely track crop phenology development \citep{sitokonstantinou2020sentinel, sitokonstantinou2023fuzzy} or identify any abrupt changes (e.g., cultivation practices, natural disasters etc.) \citep{keay2023automated}, crucial points may be left out, and as a result, the performance would be sub-optimal, especially when the gaps are long \citep{moreno2020}.

Equally, the combination of various multi-sensor data can result in more dense satellite image time series (e.g., fusion of PlanetScope and Sentinel-2) which could, in turn, increase the number of cloud-free observations. However, there are still challenges that need to be taken into account, such as the different scales of spatial resolution \citep{zhu2016} and the potentially high financial cost of other sources \citep{sadeh2021} that acts as a barrier to the scalability of the methodologies. Additionally, complementary inputs from ancillary optical sensors usually require numerous pre-processing steps to achieve the same data distribution \citep{claverie2018}, and the provision of a sufficient number of additional observations is at stake since they are affected by cloud coverage as well. 

Contrary to the optical systems, SAR sensors can offer imagery under any weather circumstances, making them a very important source of information for agricultural monitoring in areas with frequent clouds \citep{liu2019research}. Moreover, backscatter coefficients can detect the geometric, structural, and dielectric properties of plants \citep{mandal2020dual}, rendering SAR data suitable for tasks such as crop classification \citep{ioannidou2022} and grassland monitoring \citep{voormansik2016}.
Overall, a dynamic combination of SAR with the available optical data can i) enrich the feature space and enhance the overall performance of downstream tasks and ii) fill in the missing values in optical products \citep{garioud2021}.

In this study, we utilize Sentinel-1 and Sentinel-2 time series data to optimize the filling of missing NDVI values. Our approach involves a deep learning architecture that combines Convolutional and Recurrent Neural Networks (CNN-RNN). Specifically, we integrate Sentinel-1 backscatter and coherence time series with the available Sentinel-2 NDVI images to construct continuous NDVI time series of a fixed 6-day temporal resolution, focusing on grasslands across Lithuania.
Then, we evaluate the impact of our generated continuous grassland time series in the downstream task of detecting grassland mowing activity. We choose the CNN-RNN architecture as it has been applied in a similar study of \cite{zhao2020}.
The main contributions of this study are summarized as follows:

\begin{enumerate}
\item Developed a CNN-RNN model that integrates Sentinel-1 and the cloud-free Sentinel-2 NDVI observations in order to construct continuous NDVI time series of a 6-day temporal step at the pixel level.
\item Assessed the performance of our cloud gap-filling method for different cloud coverage scenarios. Additionally, we performed a comparison analysis with other widely-used baseline interpolation methods. 
\item Evaluated the added value of utilizing the continuous NDVI time series on detecting grassland mowing events.
\end{enumerate}

\section{Related work}

The reconstruction of optical images has been widely studied, with most papers focusing on spectral, temporal, or spectral-temporal methods. Among these, spectral-temporal methods offer the best performance \citep{shen2015missing}. Recently, there has been an increase in studies that specifically target NDVI reconstruction. The methods used by these studies can be divided into three main groups: i) temporal-based, ii) frequency-based, and iii) hybrid \citep{li2021high}. When it comes to vegetation monitoring, the most common are the temporal-based methods \citep{Yang2022}. These methods exploit the phenological continuity of NDVI time series, but they lack precision when long temporal gaps are present. Frequency-based methods \citep{roerink2000} have low computational complexity and are easy to implement, however, they are not able to handle sudden changes in vegetation growth, such as mowing events. On the other hand, hybrid methods \citep{chu2021,zhao2023}, which utilize both temporal and spatial information, can achieve better results and are more capable of tackling the aforementioned issues. 
In the publication of \cite{julien2019optimizing}, the authors evaluated and explored different parameterizations of five temporal reconstruction techniques. Even though the performance of some models was quite promising, the deviations in cloud-prone areas remained high. 

The abovementioned methods focus on filling in missing NDVI values using only the available cloud-free time series. However, utilizing ancillary data sources, such as SAR data, provides valuable information that can be correlated with optical-derived vegetation indices.
In recent years, the extensive adoption of Artificial Intelligence (AI) has led to a rise in the utilization of Machine Learning (ML) approaches for gap-filling \citep{wang2019,garioud2021}. These approaches have shown promising performance across various land covers, including crops, grasslands, and forests. Unlike empirical models, which rely solely on initial hyperparameters and thresholds, these fully data-driven approaches require a large number of annotated training and validation samples to achieve high performance and ensure transferability and generalization.

In literature, one can find many studies on the joint exploitation of Sentinel-1 and Sentinel-2 observations \citep{veloso2017,wang2017,scarpa2018}. For instance, standard ML regression routines have been applied to capture the correlations between the different sources and generate optical data, based exclusively on SAR observations \citep{mohite2020}. Additionally, \citep{pinto2022} demonstrated the effectiveness of random forest in combining Sentinel-1 and Landsat-8 satellite data to predict vegetation indices accurately. Similarly, \citep{pipia2019} explored Gaussian processes to model the relationship between SAR and optical data, emphasizing the method’s robustness in filling cloud-induced data gaps.
However, these approaches do not consider the temporal dimension of the input data and have proven to be very sensitive to the initial hyperparameter selection, depending on the type of vegetation cover and area characteristics, which challenges their generalization and overall effectiveness.

The recent advancement in Deep Learning (DL), combined with increased computational capacities and the availability of open-access large datasets, paved the way for the successful adoption of fully automated data-driven approaches. Several works have been published that use CNN or Generative Adversarial Neural Networks (GANs) to exploit spatial characteristics and transform single-date SAR images into equivalent optical ones \citep{reyes2019,meraner2020,rossberg2023globally}. However, these approaches do not consider the dimension of time either, which is valuable information for crop monitoring, and usually have high computational complexity. On the other hand, RNNs are considered ideal for time series estimation because they are able to track temporal dependencies. Several architectures have been developed to exploit spatio-temporal data and enhance the prediction performance \citep{ienco2019,Ebel_and_Garnot_2023_CVPR}.
Interestingly, \cite{rusvurm2018} found that certain RNN cells evidenced sensitivity to cloudy pixels, potentially acting as an internal filtering mechanism.

Over the past several years, deep learning techniques that incorporate the temporal dimension have been effectively used for reconstructing optical vegetation indices using only SAR data. In \cite{zhao2020}, the authors proposed the MCNN-seq framework, combining CNN and Long Short-Term Memory (LSTM) layers with an attention mechanism to capture long-term dependencies in the time series. However, its generalization capability needs further assessment due to the small number of crop classes and the limited geographical extent (i.e., Southern California).
In \cite{garioud2021} the authors introduced SenRVM, a deep RNN design that processes Sentinel-1 SAR data alongside auxiliary information such as topographic and climatic details. This approach showed promising performance across various vegetation types in France, but scaling up to a national level poses challenges due to the extensive data requirements and computational complexity.
No less important, \cite{li2022} applied a Transformer Temporal-Spatial Model (TTSM) on a larger spatial scale (i.e., 1M pixels), outperforming other deep learning architectures. They used transfer learning to test their model on neighboring areas in China, achieving promising results for crops and forests. However, the geographical proximity of the test regions suggests limited generalizability.
Lately, \cite{rovberg_and_schmitt_2024} proposed a novel methodology that combines sparse optical NDVI time series with denser but less accurate SAR-derived NDVI time series using a Gated Recurrent Unit (GRU), applied on 1206 regions worldwide between 2019 and 2021.
Overall, the aforementioned solutions focus on the general problem of reconstructing NDVI from SAR data across various land cover types. Yet, to the best of our knowledge, no studies have explicitly addressed the impact of cloudy images on the detection of mowing events in grasslands.

Grassland monitoring with remote sensing data has been widely studied. In order to detect grassland mowing, optical indices have been linked with interferometric SAR coherence of X and C bands \citep{ali2017,devroey2021}. However, confounding factors such as grazing events or other farming activities, land factors and meteorological conditions can affect the accuracy of the results \citep{tamm2016}. To this purpose, combining optical and radar data can enable the tracking of sudden events and mapping of grasslands use intensity \citep{reinermann2022detection}.

Current methods using fixed thresholds can lead to performance instabilities \citep{garioud2021}. 
The rule-based Sen4CAP algorithm \citep{devroey2022} includes a Sentinel-2 component in its processing chain for mowing detection, which searches for sharp drops between consecutive NDVI cloud-free observations, surpassing a predefined threshold. The algorithm achieved 69\% accuracy across six European sites, though individual site metrics varied significantly (up to 30\%). In 2019, it demonstrated an F1-score of 79\% over 150 parcels in the Netherlands. However, when evaluated on 118 cases in Lithuania, the F1-score dropped to 60\%, primarily due to a significantly reduced precision of 49\%.
Another widely used approach is the algorithm proposed by \cite{schwieder2021}. This method considers the residuals of the assessed vegetation index values from an interpolated time series formulated based on the seasonal peak timestamps that represent the ideal growing season curve. Indicatively, the algorithm achieved an overall F1-score of 67\% on 180 grassland parcels in Germany for 2020.

Similarly, several studies have used machine learning for mowing event detection \citep{taravat2019,komisarenko2022}. In particular, \cite{lobert2021} applied a deep learning model on Sentinel-1/2 and Landsat 8 time series data in Germany, identifying limitations such as potential event omission due to optical data interpolation to Sentinel-1 intervals. \cite{Lange2022} developed a CNN-based algorithm to quantify land-use intensity over grasslands using Sentinel-2 NDVI time series. Finally, \cite{Holtgrave2023} explored a two-step transferable approach that initially uses optical, SAR, and weather data to fill gaps in optical time series, followed by a binary classification to distinguish between mown and unmown grassland cases.

\section{Materials}\label{sec:materials}

\subsection{Study area}

Lithuania is located in Northeastern Europe and covers an area of 65,300 km². Its temperate climate is influenced by both maritime and continental factors, and it is primarily defined as humid continental under the Köppen classification scheme. However, a small coastal zone adjacent to the Baltic Sea can be considered oceanic. It is mainly characterized by cold temperatures, frequent rainfalls, and snowy winters. As a result, four climatic regions have been identified: Coastal, Samogitian, Middle Lowlands, and Southeastern Highlands \citep{lithuania_climatic_regions}. 

Grasslands are the most prevalent crop types and are spread equally across the country. 
They are divided into two main categories: permanent and temporary grasslands. Permanent are mostly perennial pastures and meadows, or other natural/semi-natural grassland parcels of the same land use for more than five years. On the other hand, temporary grasslands are the cases of newly updated grassland usage, of less than five years. In Lithuania, CAP requires farmers to perform at least one agricultural activity (mowing or grazing), usually until the 1st of August of every cultivation period.

\begin{figure}[!ht]
\centering
\includegraphics[width=\columnwidth]{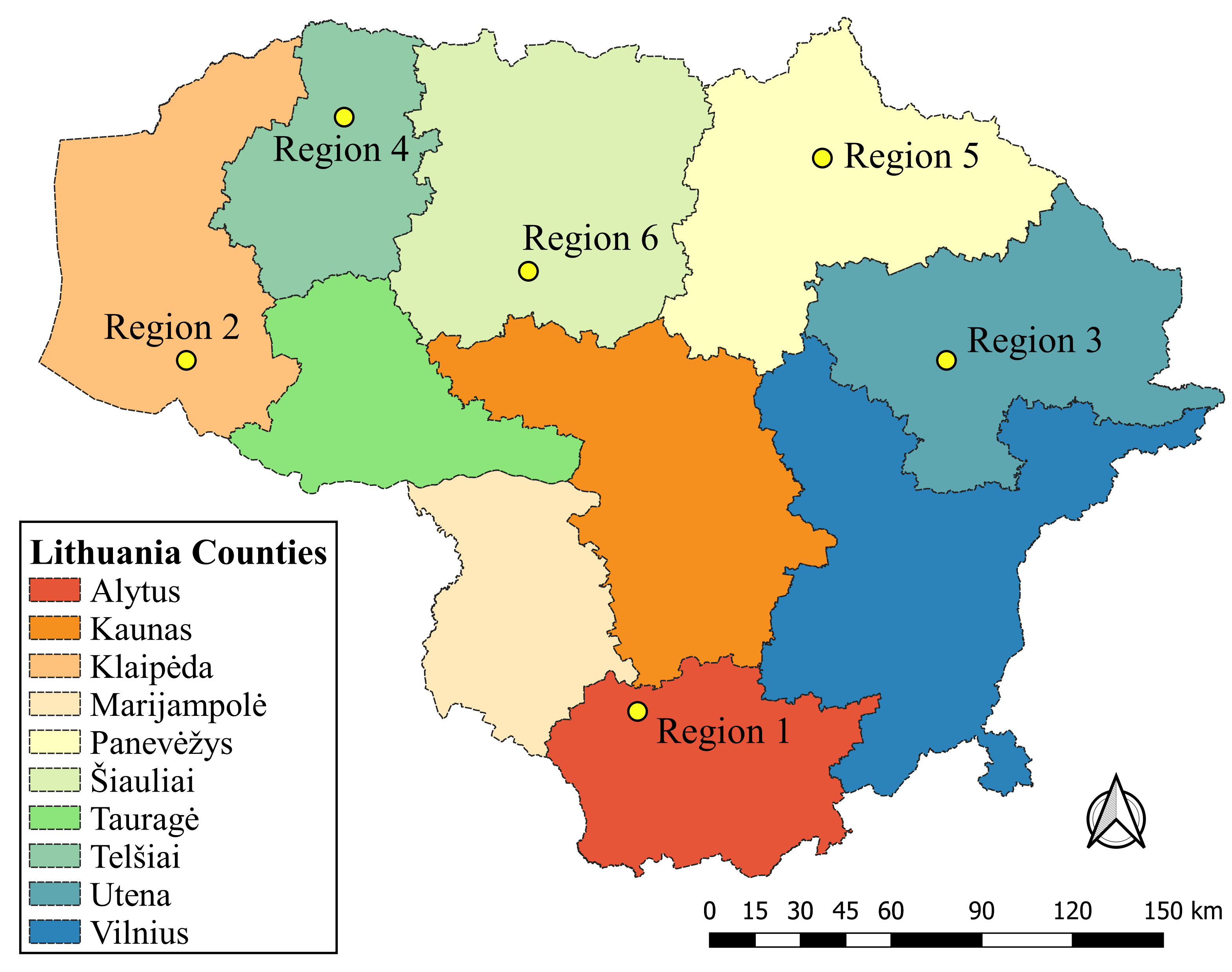}
\caption{Study regions in Lithuania.}
\label{fig:lithuania_study_sites}
\end{figure}

In this study, six distinct regions were selected based on their different agro-climatic and morphological characteristics as determined by Lithuania's climatic region map and the distribution of grassland cultivation (see Fig. \ref{fig:lithuania_study_sites}). 

Region 1 is located in the county of Alytus (southern Lithuania) at the center of the Dzūkija region which is mostly covered with pine forests. Despite being watered by the Nemunas River, this area is unsuitable for agricultural activities due to the poor quality of the soil, which is mostly sandy. Regions 3 and 5 are located around the counties of Utena and Panevėžys, respectively, in the northeast part of the country and inside the region of Aukštaitija, which contains more urban elements than the rest. In the Samogitia region, an area in northwestern Lithuania with many lakes and intense agricultural activity, we find Regions 4 and 6, near the counties of Telšiai and Šiauliai, respectively. Finally, Region 2 is in the county of Klaipėda, located in the western part of the country. Its climate is highly influenced by the Baltic Sea, with very frequent storms, as well as snowfalls and gales during winter. 
Table \ref{tab:spatial_stats} contains the number of grassland parcels and their average size for each study region. The average size of grassland parcels in the entire country is over 1 ha (see the respective Fig. S1 in \ref{sec:supplementary_data}), which is substantially bigger in comparison to other European countries. However, there are a few cases with size less than 0.1 ha, which results in mixed information within the Sentinel-2 pixels and have been eliminated from our study. 

\begin{table}[!ht]
\centering
\caption{Number of grassland parcels and their respective average size in hectares (ha) for the six study regions.}
\label{tab:spatial_stats}
\begin{adjustbox}{width=\columnwidth,center}
\begin{tabular}{@{}p{0.4\columnwidth}>{\centering\arraybackslash}p{0.3\columnwidth}>{\centering\arraybackslash}p{0.4\columnwidth}@{}}
\toprule
{Region} & {\# parcels} & {Avg. parcel size (ha)}  \\ 
\midrule
Region 1 (Alytus) & 10,426 & 1.33 \\ 
Region 2 (Klaipėda) & 7,497  & 2.35 \\ 
Region 3 (Utena) & 11,181 & 1.75 \\ 
Region 4 (Telšiai) & 9,075  & 1.82 \\ 
Region 5 (Panevėžys) & 2,760  & 2.58 \\ 
Region 6 (Šiauliai) & 4,287  & 1.80 \\
\bottomrule
\end{tabular}
\end{adjustbox}
\end{table}

\subsection{Sentinel Data}

For the present study, we utilized both Sentinel-1 and Sentinel-2 satellite imagery. We obtained Sentinel-1 Ground-Range-Detected (GRD) and Single-Look-Complex (SLC) products, as well as Sentinel-2 L1C data for the year 2020 from the CreoDIAS data repository. It is worth noting that the data acquisition period is before December 2021, prior to the instrument electronics power supply anomaly that occurred on Sentinel-1B \citep{Pinheiro2022}. As a result, the revisit time for the radar data is approximately 6 days.  

The study regions are spread evenly throughout Lithuania, with multiple orbits intersecting with each other to cover the regions of interest.
From the total of 14 available Sentinel-2 tiles, 8 (i.e., 34UEG, 34UFG, 34UFF, 34VEH, 35ULA, 35ULB, 35UMB and 35VLC) intersect with the respective study regions. Moreover, two Sentinel-1 relative orbits were selected, namely 58 and 131, which in combination cover the entire country of Lithuania and guarantee consistent acquisition geometry of the respective time series per study region. Additionally, these specific orbits were acquired in the afternoon, which is crucial for grassland monitoring since morning dew can cause an unexpected decrease of coherence on both mowed and unmowed parcels \citep{tamm2016}.

\subsubsection{Data Processing}

A time series of Sentinel-2 images from April 1st to September 30th, 2020, were atmospherically corrected with Sen2Cor \citep{sen2cor_new} to transform the Top of Atmosphere (TOA) Level 1C data into Bottom of Atmosphere (BOA) Level 2A products, also generating a Scene Classification Layer (SCL) with cloud and land feature classifications. The images were reprojected and clipped to the study areas, focusing on NDVI calculation from the B04 (red) and B08 (near-infrared) bands, both at 10 m resolution.

Sentinel-1 Level-1 GRD data in Interferometric Wide (IW) swath mode and SLC products were used to derive backscatter ($\mathrm{\sigma_{0}}$) and interferometric coherence coefficients, respectively. GRD products pre-processing with the snappy library included radiometric calibration, speckle filtering with the Refined Lee Filter (RLF), terrain correction using Shuttle Radar Topography Mission (SRTM) 10-m, and conversion of backscatter coefficients into decibels (dB).
Selected bursts using the TOPS SPLIT function from SLC products were co-registered using Back-Geocoding with updated orbit information and a Digital Elevation Model (DEM) to align split images, allowing for coherence estimation between reference and secondary images. Finally, backscatter and coherence coefficients were generated in VV and VH polarization modes, along with derived products including mixed coherence, Radar Vegetation Index (RVI), backscatter ratio, and cross-ratio.

In Table \ref{tab:features}, we summarize the list of all features extracted from Sentinel-1 and Sentinel-2 for the development of this study.

\begin{table}[!ht]
\centering
\caption{Features used in this study and their relevant description or formula. $\mathrm{\sigma_0}$ refers to the backscatter coefficient and \textit{coh} is an abbreviation for coherence.}
\label{tab:features}
\begin{adjustbox}{width=\columnwidth,center}
\begin{tabular}{@{}p{0.4\columnwidth}p{0.65\columnwidth}@{}}
\toprule
Feature & Description \\
\midrule
\rule{0pt}{2ex}   $\mathrm{\sigma_0}$ & VV and VH bands from Sentinel-1 GRD \\
\rule{0pt}{2ex}   coherence & VV and VH bands from Sentinel-1 SLC \\
\rule{0pt}{2ex}   $\mathrm{\sigma_0}$ ratio & $\mathrm{\displaystyle\frac{\sigma_{0} VV }{\sigma_{0} VH}}$ \\
\rule{0pt}{2ex}   $\mathrm{\sigma_0}$ cross-ratio & $\mathrm{\displaystyle\sigma_{0} VH - \sigma_{0} VV}$ \\
\rule{0pt}{2ex}   mixed coherence & $\mathrm{\displaystyle\sqrt{coh_{VV} \cdot coh_{VH}}}$ \\
\rule{0pt}{4ex}   RVI & $\mathrm{\displaystyle\frac{4\cdot\sigma_{0} VH}{\sigma_{0} VH + \sigma_{0} VV}}$ \\
\rule{0pt}{4ex}   NDVI & $\mathrm{\displaystyle\frac{NIR-RED}{NIR+RED}}$ \\
\bottomrule
\end{tabular}
\end{adjustbox}
\end{table}

To handle the large amount of data efficiently, we utilized the Agriculture monitoring Data Cube (ADC) developed by \cite{datacube_2022}. To ensure temporal consistency across regions, we established a common framework based on acquisition dates from Sentinel-1’s relative orbit 131 as the primary reference. This approach aligned Sentinel-1 and Sentinel-2 data to the closest dates on a unified 6-day interval from April 9 to September 24, 2020, resulting in 29 consistent timestamps across the study period. Cloud-free Sentinel-2 timestamps were shifted to match the nearest reference date if within a 3-day window; otherwise, gaps were marked as empty (i.e., with NaN values).

\begin{figure}[!ht]
\centering
\includegraphics[width=0.8\columnwidth]{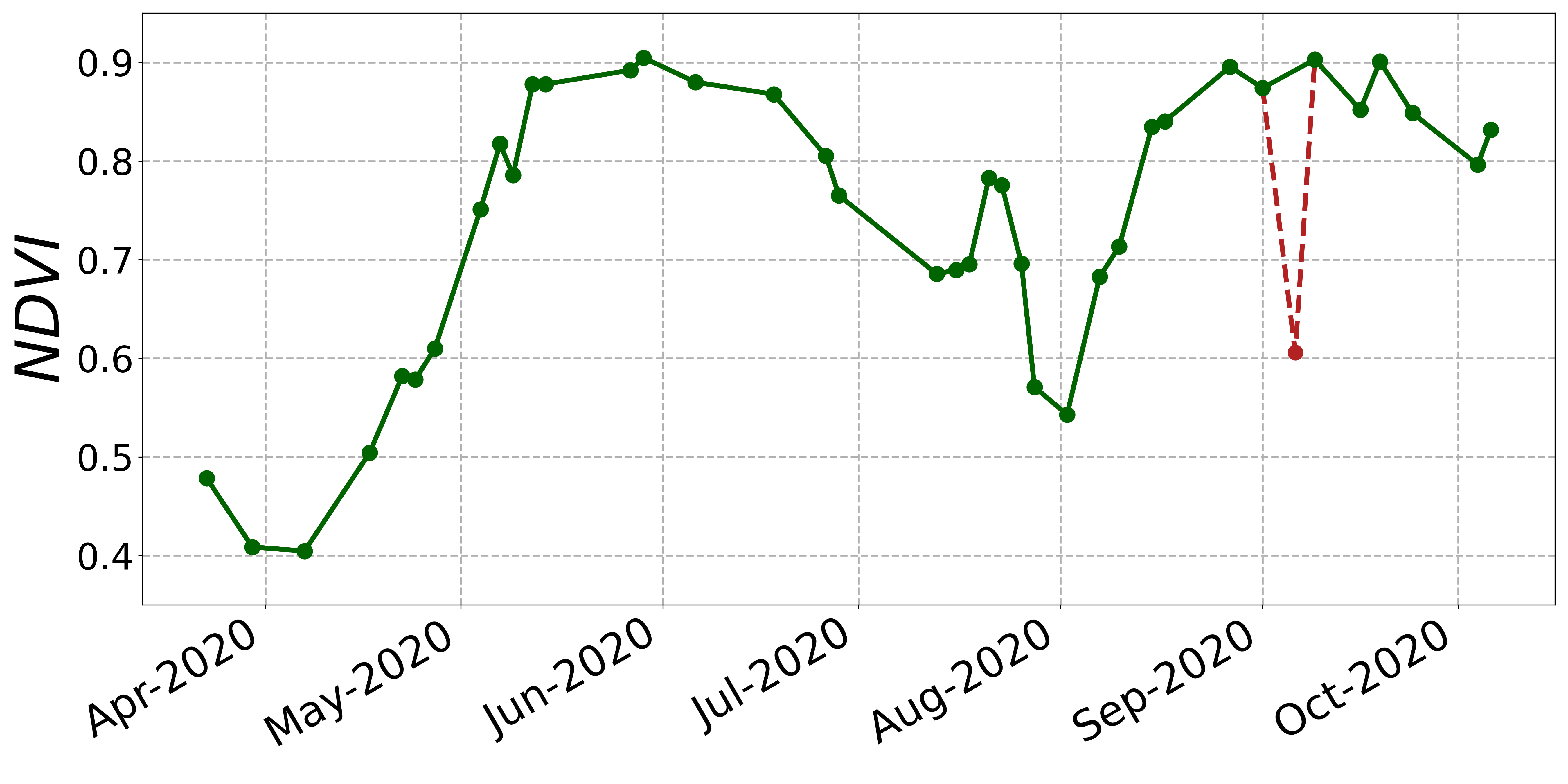}
\caption{Example of NDVI time series outlier removal. The red dashed line shows the raw NDVI time series (i.e., including noise), and the green line illustrates the NDVI after filtering potential cloudy measurements remaining from the initial cloud masking step.}
\label{fig:ts_outlier_removal}
\end{figure}

Missing values resulted from noisy pixels that had been masked out due to clouds from the SCL product of the sen2Cor software. However, these cloud masks are not always accurate, especially in cases of cloud shadows, thin clouds, or pixels adjacent to clouds, which cause an underestimation of the NDVI value that can potentially lead to incorrect detection of mowing events \citep{tarrio2020}. 
To address this issue, we implemented a threshold-based outlier removal algorithm (Algorithm \ref{alg:outliers_removal}) to filter NDVI values with sudden drops (threshold $\mathrm{\alpha}$) followed by a quick recovery (threshold $\mathrm{\beta}$) between two successive steps (Fig. \ref{fig:ts_outlier_removal}). In our case, we configured both  $\mathrm{\alpha}$ and  $\mathrm{\beta}$ threshold parameters to 0.15.
This deliberate choice of a relatively high threshold, minimizes the overall number of cases that are removed, ensuring that only certain noisy timestamps are filtered out.
Additionally, to prevent filtering cases of gradual NDVI decline due to grassland senescence, an additional threshold $\mathrm{\gamma = -0.05}$ was used to compare neighboring values.
An equivalent approach has been developed by \cite{komisarenko2022} using NDVI triplets and the rhombus equation for the identification of outlier cases. Similarly, in the recent study of \cite{WATZIG2023113577} a gradient boosting algorithm was trained to detect false positive mowing predictions caused by missed clouds.

\begin{algorithm}[!ht]
\small
\caption{Identification and removal of outlier cases}
\label{alg:outliers_removal}
\begin{algorithmic}
\Procedure{Outliers Detection}{$\alpha,\beta,\gamma$}       
    \State Read series of raw cloud-free NDVI values: $S$ 
    \State $T=len(S)$
    \For{\texttt{$t \in T$}}
        \vspace{4pt}
        \State $C_{\alpha} \rightarrow \displaystyle \frac{S(t)-S(t-1)}{DoY(t)-DoY(t-1)}$
        \vspace{4pt}
        \State $C_{\beta} \rightarrow \displaystyle \frac{S(t+1)-S(t)}{DoY(t+1)-DoY(t)}$
        \vspace{4pt}
        \State $C_{\gamma} \displaystyle \rightarrow S(t+1)-S(t-1)$
        \vspace{3pt}
        \If{($C_{\alpha} \geq \alpha$ \textbf{and} $C_{\beta} \geq \beta$ \textbf{and} $C_{\gamma} \geq \gamma$)}
            \State $S_{t} \rightarrow$ \textbf{NaN}
        \EndIf
    \EndFor
\EndProcedure
\end{algorithmic}
\end{algorithm}

\subsubsection{Annotated data}\label{sec:photointerpretation}
One of the main objectives of this study is to enhance event detection algorithms by utilizing artificially generated NDVI time series.
However, in order to evaluate the effectiveness of these algorithms, labeled data are necessary, which are typically collected through expensive and time-consuming in-situ campaigns. As conducting such campaigns on a large scale is not feasible, we employed a photo-interpretation process, carried out by three experts, to annotate data for our study.

To streamline the assessment process, we utilized validation data from the Lithuanian Paying Agency (NPA) as a starting point within the regions of interest. The parcels selected were compliant with national mowing regulations, meaning that at least one agricultural event had occurred by August 1st, although the exact timing of this event was unknown. Moreover, an equal number of random samples were included. For the photo-interpretation process, we selected parcels with minimal cloud coverage during the inference period. This approach minimized the risk of overlooking cases that might have gone undetected due to persistent cloud cover.

\begin{figure}[!ht]
\centering
\includegraphics[width=0.8\columnwidth]{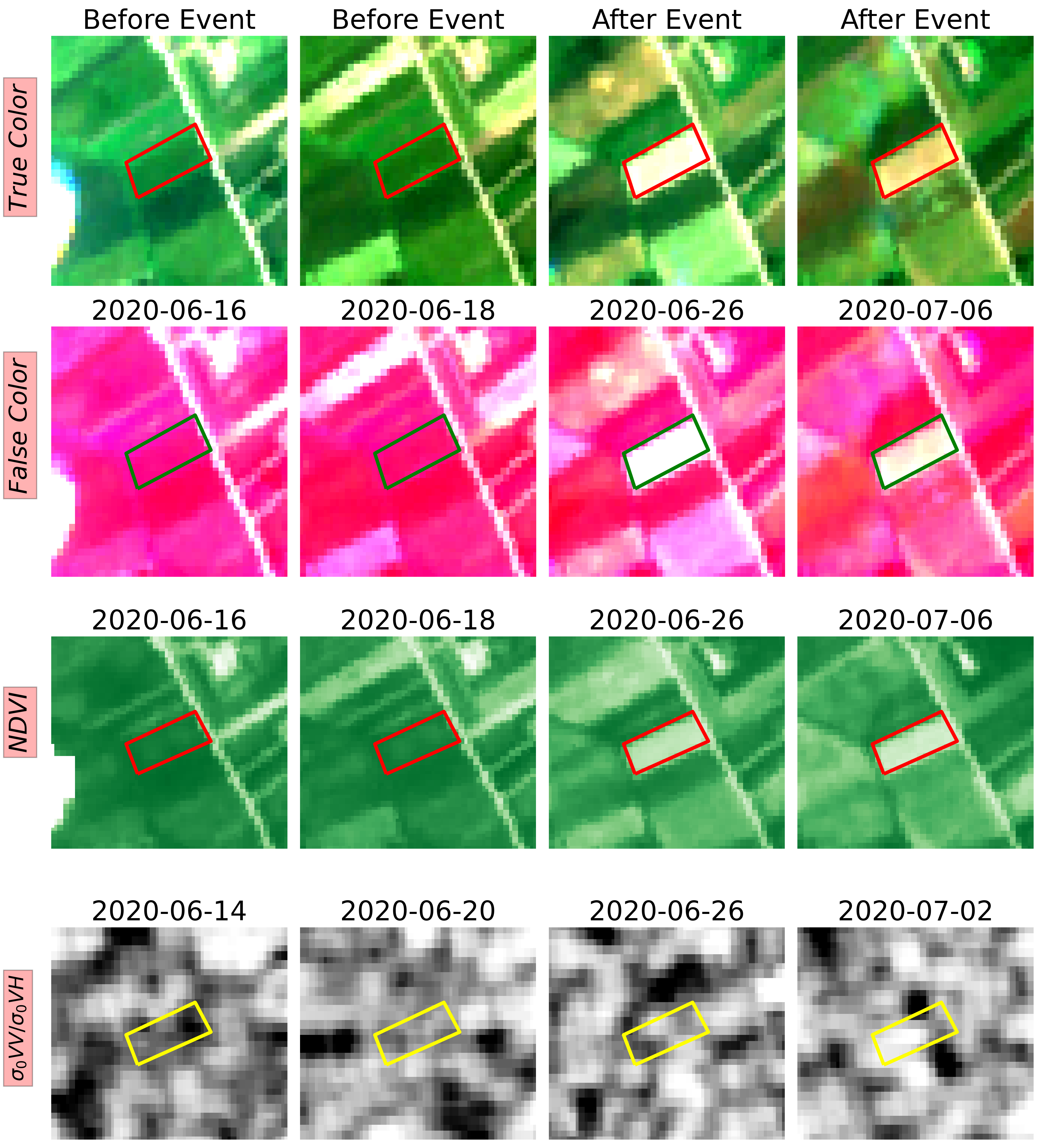}
\caption{Time series of Sentinel-1 and Sentinel-2 images for detecting a grassland mowing event. An event can be observed between June 18th and June 26th. The integration of Sentinel-1 imagery further narrows down the detection period of the event.}
\label{fig:mowing_photo_interp}
\end{figure}

The photo-interpretation process was carried out by three experts who used multi-temporal Sentinel-1 and Sentinel-2 images as reference. Since Sentinel satellites do not offer daily observations, the labeling process involved identifying the closest timestamps before and after a grassland event took place, from early April to late September. Each expert independently assigned labels on the acquisition dates of Sentinel-2, indicating the date before and after a grassland event occurred. The samples were distributed evenly across the six regions to ensure reliable model optimization and results assessment. Since SAR coefficients are sensitive to changes in the canopy of grasslands, the photo-interpretation analysis additionally incorporated Sentinel-1 data. Steep changes between successive images may showcase the occurrence of a potential grassland event. For this reason, true color (B04-B03-B02) and false color (B08-B04-B03) composites of Sentinel-2 satellite image time series are particularly useful in identifying the presence of an event (Fig. \ref{fig:mowing_photo_interp}).

Then, the examination of NDVI and SAR multi-temporal signatures was used to guide the decision of the photo-interpreter. Their decision was dictated by the fact that NDVI gets much lower values after an event (at least a decrease of 10\%), while SAR coefficients act the opposite (Rules 1-2). Secondly, an event must take place in compact dense areas and not scattered spots inside grassland parcels (Rule 3). If all the above are met, the final decision lies on the consequent values of NDVI (Rule 4). Cases where NDVI does not increase shortly (usually up to 15 days) after the potential event are excluded and are not considered valid events. This behavior many times resembles harvest events, and thus such cases may have been wrongly characterized as grasslands.

Following the visual assessments by the experts, final labels were determined through a majority voting system. Specifically, when two experts agreed on a label, this was considered the correct one. The agreement percentage between Expert 1 and 2 was 88\%, between Expert 1 and 3 was 93\%, and between Expert 2 and 3 was 86\%. However, in around 5\% of all cases, a disagreement emerged among all three experts, necessitating a collective decision to determine the final label. 

The procedure was applied in all six study regions and the final dataset includes 803 parcels, which correspond to 287,281 individual pixels (see Table \ref{tab:mowing_region}). In this dataset, 650 cases had one mowing event, 99 cases had multiple mowing events, and 54 cases showed no evidence of mowing. Notably, a significant amount of the identified mowing events occurred during June and early July. Further analysis of the NDVI profiles revealed that larger drops between consecutive NDVI values were more prevalent in June and July, whereas August exhibited smaller differences, complicating the detection of mowing events. Moreover, the distribution of NDVI differences concerning the number of mowing events indicated significantly higher values for cases with a single mowing event. Conversely, cases with two mowing events showed smaller NDVI differences, especially for the second mowing event, suggesting a reduced impact on the NDVI profile (refer to Fig. S8 and S9 in the \ref{sec:supplementary_data} for more details). This may be attributed to the prevalence of grazing practices in Lithuania \citep{devroey2022}.

\begin{table}[!ht]
\centering
\caption{Number of parcels and pixels annotated for mowing events for each region.}
\label{tab:mowing_region}
\begin{adjustbox}{width=\columnwidth,center}
\begin{tabular}{@{}p{0.4\columnwidth}>{\centering\arraybackslash}p{0.3\columnwidth}>{\centering\arraybackslash}p{0.4\columnwidth}@{}}
\toprule
{Region} & {\# parcels} & {\# pixels}  \\ 
\midrule
Region 1 (Alytus) & 274 & 64,141 \\
Region 2 (Klaipėda) & 189 &  106,556  \\
Region 3 (Utena) & 98 & 33,065 \\
Region 4 (Telšiai) & 91  & 39,254  \\
Region 5 (Panevėžys) & 79 & 23,639  \\
Region 6 (Šiauliai) & 72  & 20,626  \\
\bottomrule
\textbf{Total}  & \textbf{803}  &  \textbf{287,281} \\
\hline
\end{tabular}
\end{adjustbox}
\end{table}

\section{Methods}\label{sec:methods}

\subsection{Cloud coverage distribution }\label{sec:sparsity_analysis}

\begin{figure}[!ht]
\centering
\includegraphics[width=0.9\columnwidth]{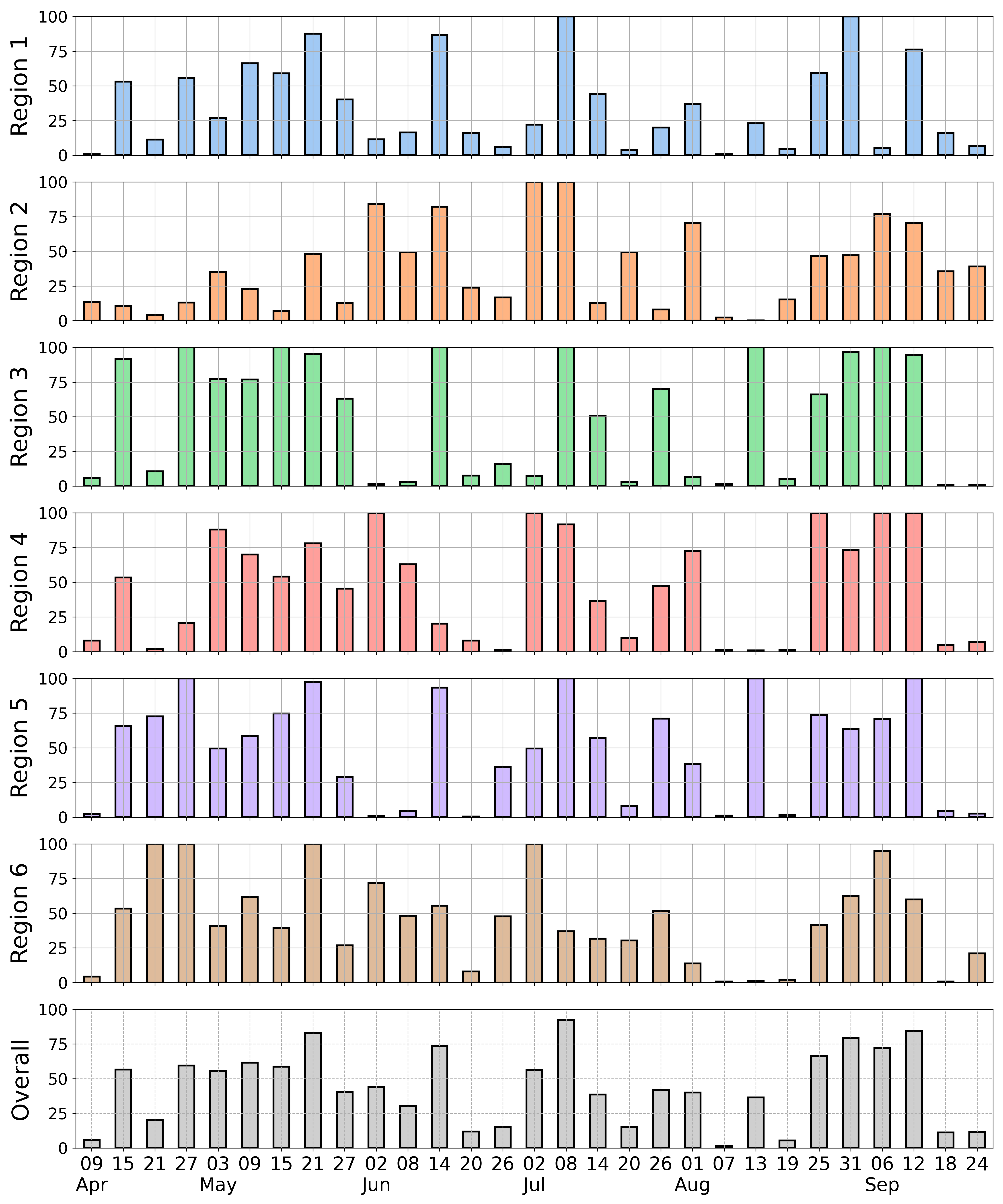}
\caption{The percentage of grassland parcels affected by clouds on each reference date, for each region and overall.}
\label{fig:sparsity per region}
\end{figure}

The study regions are characterized by unique cloud coverage profiles. 
Fig. \ref{fig:sparsity per region} visualizes the percentage of grassland parcels affected by clouds on each date and each study region (see also the respective Fig. S2 in \ref{sec:supplementary_data}). We observe significantly lower cloud coverage during the summer months (from June to August) when most mowing events occur. However, there are still cases of grassland activity that are not observed or not identified soon enough due to clouds.

We determined the cloud coverage for each grassland parcel by calculating the ratio of cloudy Sentinel-2 observations to the total acquisition dates. The mean cloud coverage for the entire dataset is approximately 45\%. To establish higher-quality ground truth information for the experiments, we worked with pixel time series and selected only the densest ones.
Specifically, we excluded cases that had i) more than one successive missing value during June and July, ii) more than 3 missing values in total during June and July, iii) more than two consecutive missing values during the rest of the assessed period, and iv) an overall cloud coverage higher than 35\%. This approach helped us ensure that we did not omit any mowing events. Finally, given that in dense time series simple temporal interpolation methodologies can effectively fill the short-term gaps, we used the Akima method \citep{akima1970} to generate the complete target NDVI time series (5339 parcels in total) served as ground truth.

\subsection{Artificial masking and NDVI input component}\label{sec:masks}

\begin{figure}[!ht]
\centering
\includegraphics[width=0.75\columnwidth]{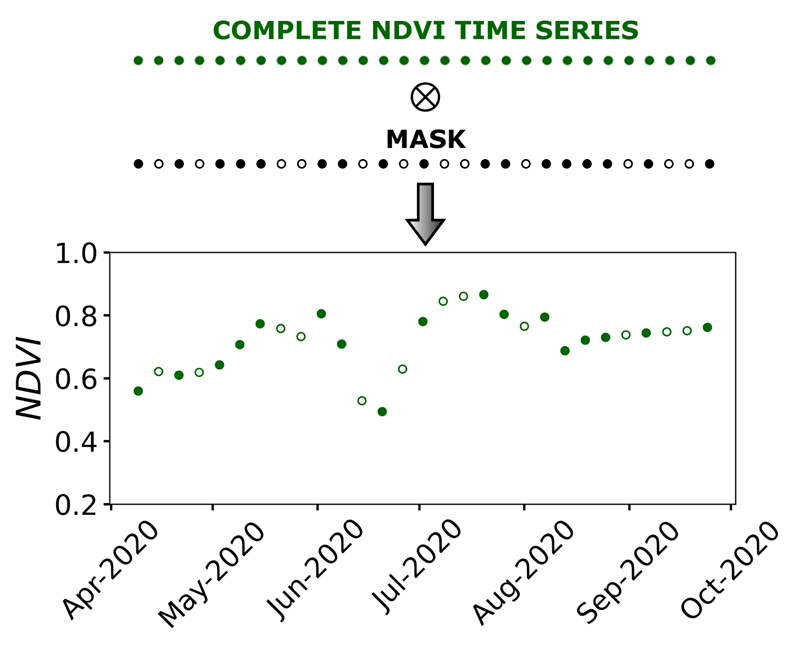}
\caption{NDVI time series with artificial cloud masking. The green dots on top of the plot show the complete NDVI time series for pixel post-processing, where the black and white dots beneath represent a simulated cloud mask. In the plot, the NDVI time series after the application of the artificial mask is displayed. Non-filled green dots indicate values that have been masked and are treated as missing data (these were uniformly substituted with a standardized placeholder value before being inserted as input into the model), while the filled green dots denote the partially available data.} 
\label{fig:masks_derivation}
\end{figure}

To train and validate our methodology, we employed artificially generated cloud masks to simulate real-world cloud coverage conditions in the NDVI time series data used as input. These cloud masks were applied to strategically obscure specific timestamps in the complete NDVI time series. For each region, we collected a set of binary time series indicating the presence or absence of clouds, based on the actual cloud coverage distribution in the corresponding area (see Section \ref{sec:sparsity_analysis}). Subsequently, we used a bootstrapping procedure to assign a random cloud mask to each NDVI time series of the same region. 
Fig. S3 (refer to section \ref{sec:supplementary_data}) illustrates the comparison between the actual and artificial cloud coverage distributions for each region.
Further, we maintained parcel consistency by applying the same mask to all pixels within the same field boundaries. Therefore, this approach ensured standardization across various parcels and maintained the alignment with the typical cloud coverage distribution in each region. An example of the NDVI values of a pixel after applying a cloud mask is shown in Fig. \ref{fig:masks_derivation}. Finally, input NDVI values affected by the generated cloud masks (i.e., artificial missing data) were uniformly replaced with a standardized placeholder value (e.g., -100) before being fed into the model. 

\subsection{Model training and architecture}

\begin{figure*}[!ht]
\centering
\includegraphics[width=\textwidth]{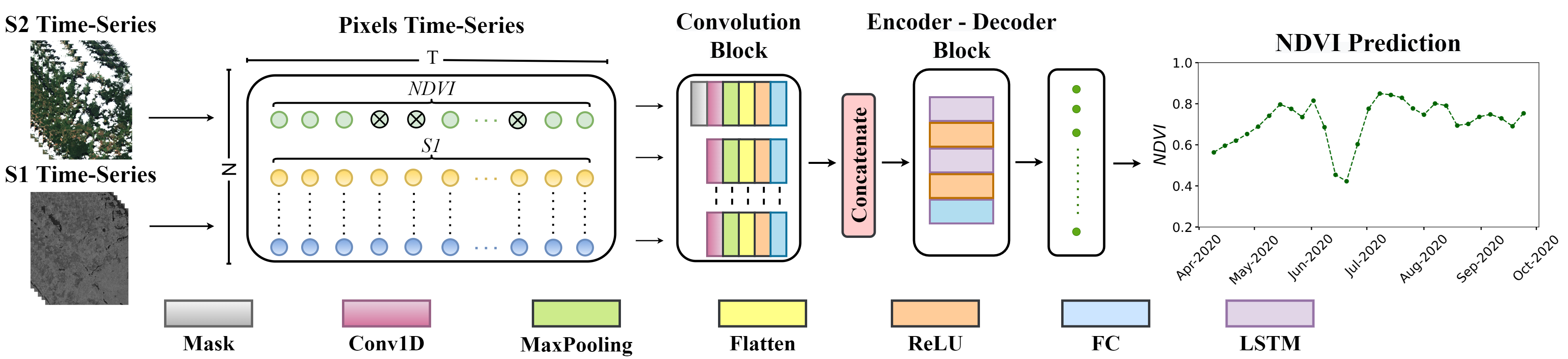}
\caption{An overview of the model architecture used for reconstructing continuous NDVI time series with a fixed 6-day interval, utilizing both SAR data and available NDVI values.}
\label{fig:model_architecture}
\end{figure*}

The proposed deep learning Sentinel-1/2 Fusion (SF) methodology is composed of sequential CNN and LSTM blocks (Fig. \ref{fig:model_architecture}). 
It is a supervised-learning neural network architecture that aims to extract full NDVI time series from Sentinel-1 time series and the available cloud-free Sentinel-2 observations. This architecture is based on the one proposed by \cite{zhao2020}, with the key modification being the inclusion of masked NDVI time series in the input feature space.

\subsubsection{CNN block}

The CNN block (CB) reduces the impact of noise from the input data and extracts meaningful features for the subsequent layers \citep{zhao2020}. 
The block consists of N parallels and independent between them, one-dimensional convolutional branches (Conv1D), where N is the number of input channels. Each Conv1D branch is followed by max-pooling layers to reduce the dimensionality of the output. Following this, a non-linear Rectified Linear Unit (ReLU) activation function is applied to convert negative values to zeros. The resulting feature maps are passed to a fully connected layer to provide the final embeddings into the subsequent LSTM blocks. 
In the case of the NDVI branch, a masking layer is added as the first layer to filter out masked NDVI inputs (e.g., -100), ensuring that the model treats cloudy data points as hidden. This enables the network to focus on reconstructing missing NDVI values based on the available cloud-free observations and SAR data, effectively handling the sparsity introduced by cloud cover.
The CNN block extracts temporal features from both the NDVI and SAR data, producing feature maps that capture various aspects of the input sequence. These maps are concatenated along the channel dimension before being passed into the LSTM encoder-decoder block, which handles the temporal dependencies in the data.

\subsubsection{Encoder-decoder block}

The encoder-decoder block (EDB), equipped with Bi-directional LSTM (BiLSTM) modules, is key to modeling the temporal dependencies within the input and the target NDVI time series. An LSTM is a type of RNN designed to capture both short-term and long-term dependencies by maintaining an internal memory state across time steps. This makes LSTMs particularly well-suited for sequential data, where past observations influence future predictions.
In the EDB, the BiLSTM modules process input sequences in both forward and backward directions, enabling the model to capture context from both past and future time steps. The encoder part processes the input embedding sequence from the CB to generate hidden states, while the decoder uses these hidden states to generate the final predicted NDVI time series without missing values.
The temporal dependencies between consecutive time steps are modeled as follows:
\begin{equation}
f_t = \sigma(W_f [h_{t-1}, x_t] + b_f) \
\end{equation}
\begin{equation}
i_t = \sigma(W_i [h_{t-1}, x_t] + b_i) \
\end{equation}
\begin{equation}
\tilde{C_t} = \tanh(W_C [h_{t-1}, x_t] + b_C) \
\end{equation}
\begin{equation}
C_t = f_t * C_{t-1} + i_t * \tilde{C_t} \
\end{equation}
\begin{equation}
o_t = \sigma(W_o [h_{t-1}, x_t] + b_o) \
\end{equation}
\begin{equation}
h_t = o_t * \tanh(C_t) \
\end{equation}
\noindent where $\mathrm{x_t}$ is the input at time $\mathrm{t}$, $\mathrm{h_{t-1}}$ is the previous hidden state, $\mathrm{c_{t-1}}$ is the previous cell state, $\mathrm{i_t}$ is the input gate, $f_t$ is the forget gate, $\mathrm{o_t}$ is the output gate, $\mathrm{c_t}$ is the cell state at time $\mathrm{t}$, and $\mathrm{h_t}$ is the hidden state at time $\mathrm{t}$. The weights and biases of the LSTM are represented by $\mathrm{W}$ and $\mathrm{b}$, respectively. The sigmoid function is denoted by $\mathrm{\sigma}$, and the hyperbolic tangent function by $\mathrm{\tanh}$. The LSTM's gating mechanisms (input, forget, and output gates) help manage information flow across time steps, ensuring that long-term dependencies are retained while mitigating the risk of vanishing gradients. 
The final NDVI time series is generated by passing the output of the EDB through a fully connected layer, which produces the reconstructed NDVI sequence based on the learned temporal patterns.

\subsubsection{Model training}\label{sec:model_training}

The CB includes an independent branch for each input layer. Each branch comprises two Conv1D layers with filter sizes of 8 and 16, both using a kernel size equal to 3. These are followed by a max pooling layer with a pool size of 3 (without striding), and finally, two fully connected layers, with 32 and 16 units, respectively. The EDB was constructed using LSTM units with a 16-unit latent space. To determine the optimal configuration, we conducted experiments with various parameter sets. Ultimately, we settled on a lightweight configuration with a total of 82,337 parameters.

For the training, we employed the Adam optimizer, due to its adaptive learning rates and computationally efficient handling of non-stationary data, together with the Mean Squared Error (MSE) loss function. The MSE loss function is defined as:
\begin{equation}\label{eq:mse}
MSE = \frac{1}{n} \sum_{i=1}^{n} (y_i - \hat{y}_i)^2
\end{equation}
where \( y_i \) represents the true values and \( \hat{y}_i \) represents the predicted values.
MSE is well-suited for our task because it penalizes large errors effectively that might arise from substantial discrepancies between predicted and target values. This is particularly beneficial for our scenario where the data can be noisy, containing preserved cloudy measurements.
The initial learning rate was set to 0.005, and the batch size to 256. The training procedure included early stopping, which would terminate the process if the loss function did not improve after three consecutive epochs, with a maximum of 30 epochs to prevent overfitting. 

In addition, incorporating the NDVI time series as both input and target can result in the model being prone to overfitting to those specific values \citep{garioud2021}. To address this issue and force the network to give more attention to masked timestamps, we have used temporal weighting for each input sample. Masked NDVI values were assigned a training weight parameter of $\mathrm{w_{\alpha}}$, while non-masked values were given a smaller weight parameter of $\mathrm{w_{\beta}}$. These parameters were set to be 0.75 and 0.25 respectively, since they provided the optimal results. Additionally, as we utilized Akima interpolation to produce a complete target time series, the values in timestamps that are not actual measurements and have been derived from the interpolation were assigned a weight of 0. This ensured they were not considered during the training and validation processes.

Finally, for NDVI reconstruction, we selected the training and validation pixel sets from the complete parcel time series after the artificial masking process (see Section \ref{sec:masks}). 
To avoid any bias in the subsequent evaluation of mowing event detection, we excluded from the training set all the pixels within the parcels that were annotated through the photo-interpretation (Section \ref{sec:photointerpretation}). 
Table \ref{tab:pixels_distro} shows the numbers of parcels and pixels of the training and validation sets in the six regions, for the NDVI reconstruction task. 
\begin{table}[!ht]
\centering
\caption{Number of training and validation pixels and parcels for each region.}
\label{tab:pixels_distro}
\begin{adjustbox}{width=\columnwidth,center}
\begin{tabular}{@{}p{0.4\columnwidth}>{\centering\arraybackslash}p{0.35\columnwidth}>{\centering\arraybackslash}p{0.35\columnwidth}@{}}
\toprule
{Region} & {\# training pixels \newline (\# parcels)} & {\# validation pixels \newline (\# parcels)}  \\ 
\midrule
Region 1 (Alytus) & 408,132 (2,002)     & 86,747 (367)\\
Region 2 (Klaipėda) & 184,452 (564)  &  140,242 (253)  \\
Region 3 (Utena) & 68,180 (364) & 36,715 (108)\\
Region 4 (Telšiai) & 158,419 (737)  & 47,785 (107)  \\
Region 5 (Panevėžys) & 49,377 (199) & 25,502 (86)  \\
Region 6  (Šiauliai) & 114,742 (451)  & 28,307 (101) \\
\bottomrule
\textbf{Total}  & \textbf{983,302 (4,317)}  &  \textbf{365,298 (1,022)} \\
\hline
\end{tabular}
\end{adjustbox}
\end{table}

The experiments were conducted on a system with an Intel Core i7-9700 CPU (8 cores, 3.0 GHz), 16GB of RAM, and an NVIDIA GeForce GTX 1660 GPU (6GB GDDR5 memory). On this setup, the average training time per epoch for a training dataset of 983,302 instances was approximately 31.8 seconds.

\subsection{Event detection} \label{sec:event_detection_methods}

We employed two distinct algorithms for mowing event detection, referenced as follows:

\begin{enumerate}

\item \textit{Mowing Detection Algorithm I (\textbf{MDA I})}: This algorithm, developed by \cite{devroey2022}, was implemented using pseudo-code from the Sen4CAP package description \citep{bontemps2022sen4cap}.

\item \textit{Mowing Detection Algorithm II (\textbf{MDA II})} is the algorithm proposed by \cite{schwieder2021}. The repository for their implementation was collected and cloned as a user-defined function in FORCE processing environment\footnote{\url{https://github.com/davidfrantz/force-udf/tree/main/python/ts/mowingDetection}}.

\end{enumerate}

Both these methods, while not perfect, presented quite satisfactory results in our dataset. Nonetheless, we want to assess whether using SF to generate an artificial cloud-free NDVI time series, can improve the mowing event detection accuracy. Thus, we used these methods as a baseline, since they are highly dependent on the NDVI information and can directly highlight the importance of the proposed methodology.

\textit{Deep Learning Mowing Detection Algorithm (\textbf{CNN-RNN})}: 
Apart from the fusion task, we utilized the SF architecture as an event detection algorithm. However, for the aim of this task, we kept only the NDVI branch at the input of the network, since the purpose was to evaluate the mowing detection accuracy given the complete NDVI time series provided. Furthermore, we replaced the existing activation function at the end of the output layer with a sigmoid one, to quantitatively assess the probability of an event occurring at each distinct timestamp. For the training of the model, we utilized the annotated labels that were derived through photo-interpretation, as mentioned in Section \ref{sec:photointerpretation}. For this purpose, we created binary series of 0s and 1s with a size equal to the length of the time series (in our case, 29). In this binary representation, the value 1 corresponds to the closest timestamp to the annotated mowing event. This labeling format allowed the model to learn intricate temporal patterns, enabling it to offer not just an event detection but also a probability estimation for the likelihood of an event occurring at each timestamp within the extensive range of evaluated dates. This dual-purpose application of the SF architecture showcased its versatility in addressing complex tasks beyond NDVI reconstruction, highlighting its potential as an effective event detection algorithm, too. It's worth noting that while we acknowledge the potential contribution of SAR branches to the model's accuracy, their inclusion in this model configuration would introduce an unfair advantage. Overall, our objective here is to assess and quantify the unique contributions of the same optical features through different models, ensuring a fair and robust evaluation.

\section{Results}\label{sec:results}

In this section, we evaluated the SF model's performance in cloud gap-filling. We compared its performance with other common interpolation methods, namely Linear Interpolation (LI), Akima Interpolation (AI), and Quadratic Interpolation (QI) \citep{akima1970, noor2014filling}. 
More specifically, LI estimates values between known data points by drawing straight lines, AI uses piecewise cubic polynomials to smooth the curves between points, while QI fits quadratic curves between adjacent points, providing an even smoother interpolation compared to the rest. 
Additionally, we assessed the effect of missing NDVI images on the detection of grassland events and the ability of the SF model to alleviate this problem.

\subsection{Gap-filling model performance}
This section evaluates the performance of the SF model on the six grassland regions.
For the evaluation, we used the metrics of Mean Absolute Error (MAE) (Eq. \ref{eq:mae})  and coefficient of determination ($\mathrm{R^2}$) (Eq. \ref{eq:r_coeff}), considering all available time steps with a cloud-free NDVI value. 
\begin{equation}\label{eq:mae}
MAE = \frac{1}{n}\sum_{i=1}^{n}|y_i-\hat{y_i}|
\end{equation}

\begin{equation}\label{eq:r_coeff}
 R^2 = 1 - \frac{\sum_{i=1}^{n} (y_i - \hat{y_i})^2}
{\sum_{i=1}^{n} (y_i - \overline{y})^2} 
\end{equation}
where $\mathrm{\hat{y_i}}$ is the predicted value of the 
$i$-th pixel, $\mathrm{y_i}$ is the corresponding true value for $\mathrm{n}$ total pixels and $\overline{y} = \frac{1}{n}\sum_{i=1}^{n}y_i$.

\begin{figure}[!ht]
\centering
\includegraphics[width=0.8\columnwidth]{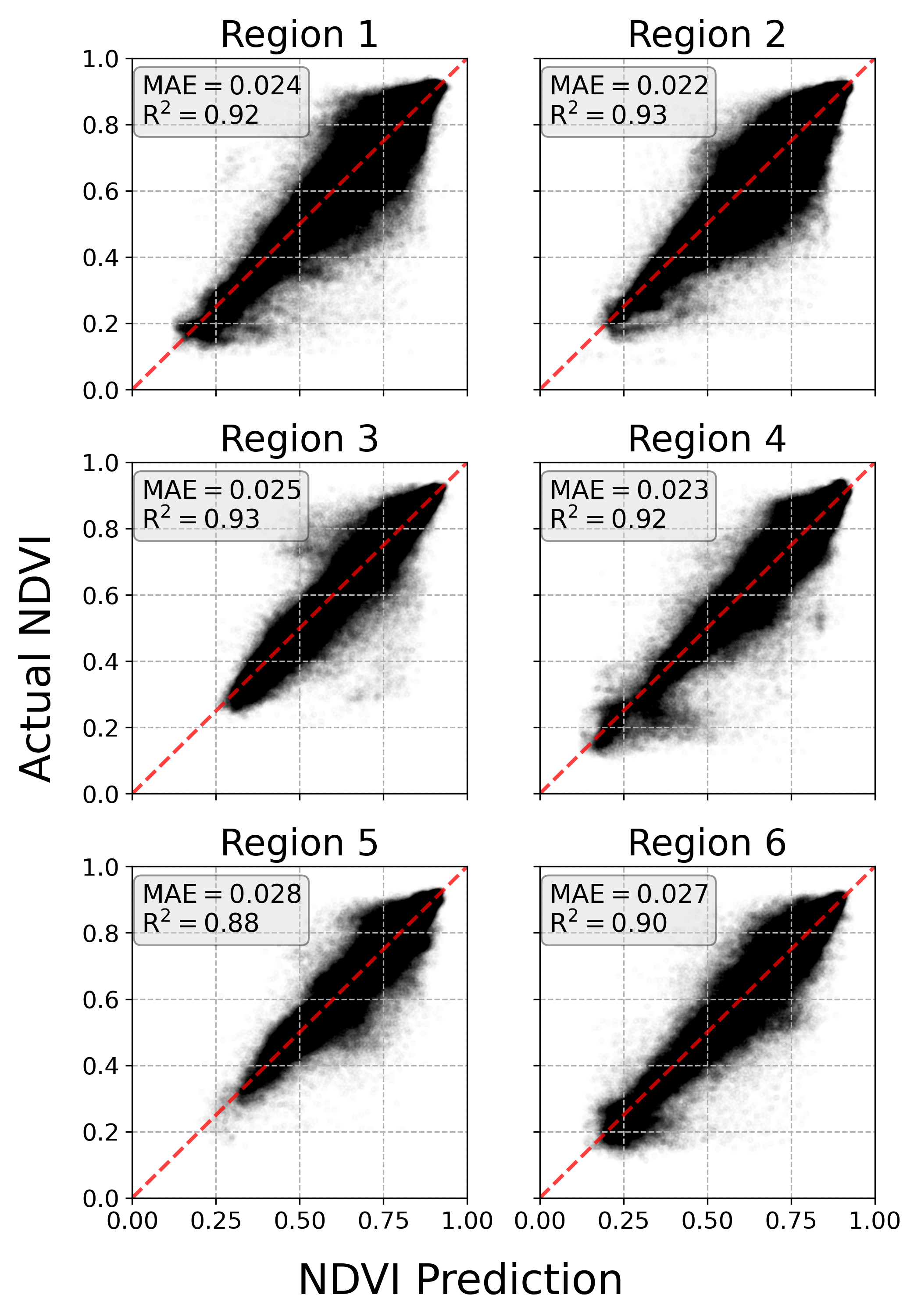}
\caption{Scatter plots between the actual NDVI values and the SF predictions for each study region.}
\label{fig:scatter_fusion}
\end{figure}

In Fig. \ref{fig:scatter_fusion}, we present scatter plots for each of the six study regions. Notably, we have noted a strong correlation between the ground truth and the predictions generated by the SF model. Region 5 stands out with relatively poorer performance, characterized by a MAE of 0.028 and a $\mathrm{R^2}$ value of 0.88. This discrepancy can be attributed to the presence of relatively high cloud coverage and a limited number of training pixels from the area. However, when assessing the other regions, we did not find any statistically significant differences in performance. Collectively, the $\mathrm{R^2}$ coefficient measures at 0.92, and the MAE is calculated at 0.024, which highlights on SF model's capacity in effectively generating missing NDVI values.

\begin{figure}[!ht]
\centering
\includegraphics[width=0.8\columnwidth]{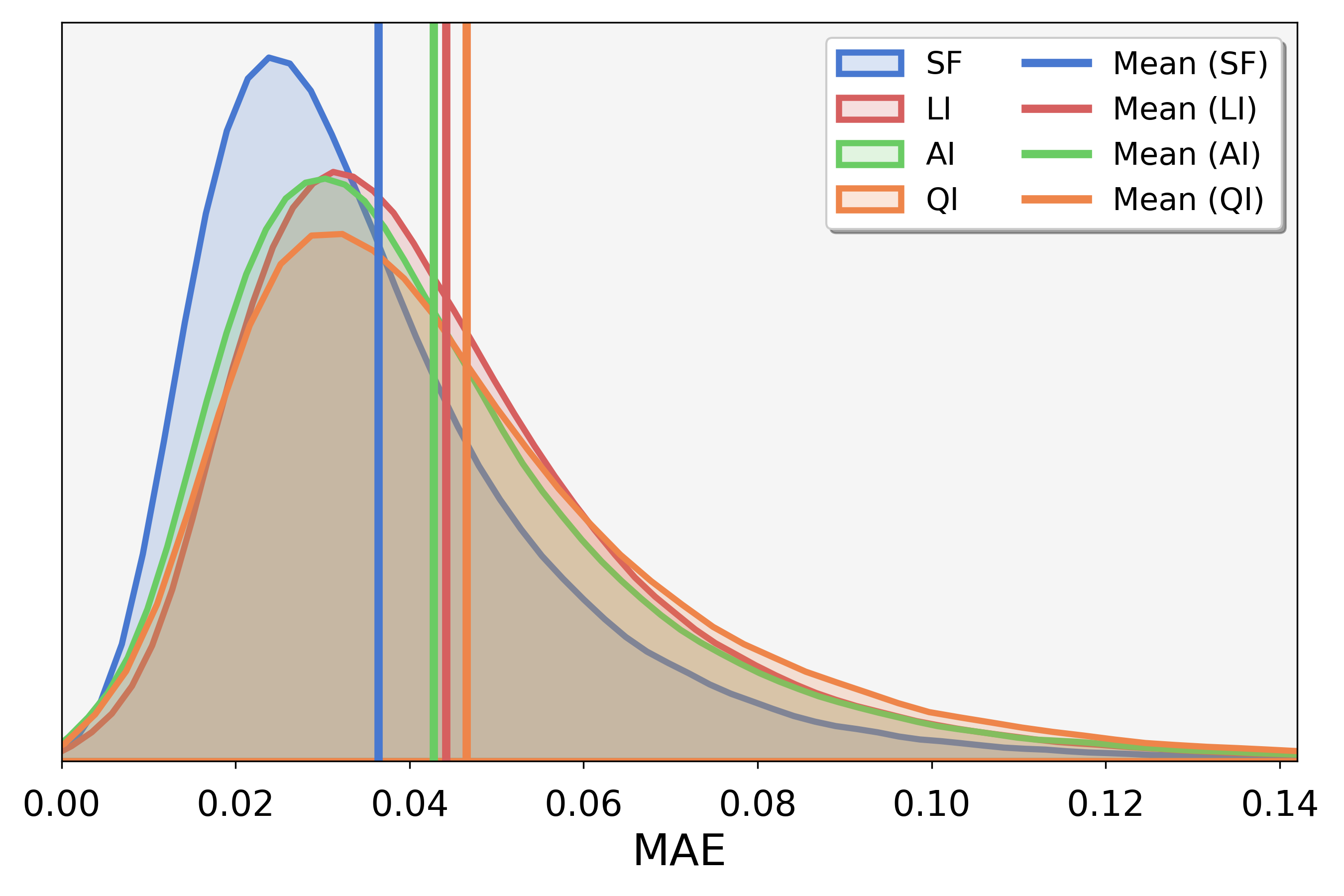}
\caption{The distribution of MAE, only on the masked timestamps, for the SF and the three interpolation methods (AI, LI, and QI).}
\label{fig:mae_distro}
\end{figure}

Fig. \ref{fig:mae_distro} presents the distribution of MAE for the SF model and the three alternative interpolation methods. This analysis focuses on timestamps that were intentionally masked in order to highlight the model's ability to fill missing NDVI values. The SF model outperformed the temporal interpolation methods, with a mean MAE of 0.036 and a median of 0.032. In contrast, the AI and LI methods exhibited less favorable results, with average MAE values of 0.043 and 0.044, respectively. The QI method performed the poorest among the approaches, with an average MAE of 0.047 and a notably higher standard deviation.

For further details regarding cloud gap-filling performance across various parameters such as study region, date of inference, cloud coverage, and length of consecutive missing values, please refer to \ref{sec:supplementary_data} Fig. S4 to S7.

\subsection{Event detection accuracy assessment} \label{sec:4_2}

In this section, we evaluate the SF methodology's contribution to the detection of grassland mowing events. We used event timestamps as labels to assess the impact of the artificially created NDVI time series on the event detection task. We examined 803 unique grassland parcels across the six study regions, which correspond to 287,281 pixels, and derived from the photo-interpretation process (Section \ref{sec:photointerpretation}).

Our evaluation investigates how the SF model affects grassland event detection by comparing two rule-based methods outlined in Section \ref{sec:event_detection_methods}.  
These methods depend exclusively on NDVI time series to identify substantial drops in their values. Consequently, they may exhibit limitations in detecting events, especially when consecutive cloudy observations are present.
They provide a time range for each event, allowing us to compare it directly with the respective range obtained from photo-interpretation. Finally, we measured the detection accuracy and the ability to predict an event correctly using recall (Eq. \ref{eq:recall}) and precision metrics (Eq. \ref{eq:precision}), and their harmonic mean using the F1-score metric (Eq. \ref{eq:f1_score}). The predictions are aggregated and evaluated at the parcel level of photo-interpretation.
\begin{equation}\label{eq:recall}
Recall = \frac{True \ Positive}{True \ Positive + False \ Negative}
\end{equation}

\begin{equation}\label{eq:precision}
 Precision = \frac{True \ Positive}{True \ Positive + False \ Positive}
\end{equation}

\begin{equation}\label{eq:f1_score}
 F_{1}-Score = \frac{2 \cdot  Precision \cdot  Recall}{Precision + Recall}
\end{equation}
A predicted event is deemed correct if its predicted date falls within twelve days (tolerance parameter) before or after the actual event, which is twice the size of the time series interval. Events falling outside this tolerance window are considered false positives. This approach helps mitigate issues arising from the 6-day interpolation used for detection. This may cause events to shift from one timestamp to another, introducing challenges when reporting results, as it can impact the accuracy and alignment of detected events with their actual occurrences. 
Furthermore, while MAE is a common metric for evaluating mowing event detection algorithms, it is not considered applicable in our case since we lack the exact date of the mowing events. Our main purpose in this study is to assess how the implementation of the SF model for reconstructing missing NDVI values enhances mowing event detection by identifying cases that would have been missed otherwise.
Nevertheless, in \ref{sec:supplementary_data} (Tables S1-S3), we present the respective results for various evaluation approaches, which highlights the independence of the SF model on the alternative criteria used to define a predicted event as correct.

Table \ref{tab:mowing_detection_performance} shows the results of MDA I and MDA II algorithms acquired from the different interpolation methods and time series without any interpolation applied. Both algorithms demonstrated similar performance, with slight variations depending on the interpolation methodology. As indicated in the respective table, all interpolation approaches exhibited a common tendency to slightly reduce recall while significantly enhancing precision accuracy for both mowing detection algorithms, compared to using unprocessed time series data. 
This can be attributed on (i) the potential events omission due to optical data interpolation to Sentinel-1 intervals \citep{lobert2021}; and (ii) the interpolation capabilities to effectively eliminate numerous abrupt changes that often arise from noisy successive NDVI values by providing consistent timestamps with a 6-day gap between them. All in all, this behavior decreases the total number of falsely detected mowing events (i.e., increased precision), reducing the number of some correctly identified cases (i.e., reduced recall). In essence, these findings highlight the compromise between precision and recall, emphasizing the role of interpolation methods in refining mowing event detection accuracy. Notably, the SF generated NDVI time series, presented the best results in both MDA I and MDA II, with the latter achieving the optimal F1-score equal to 0.84.
Eventually, it appears that the SF methodology has effectively addressed challenges associated with the inference on Sentinel-1 intervals. More specifically, for MDA I, recall values are significantly better among the interpolation methods, while for MDA II, they are improved by almost 1\% with regard to the raw time series (i.e., no interpolation applied). Furthermore, SF offered significantly higher precision (over 90\%) for both MDA I and II, with a range of 19\% to 24\% improvement over the unprocessed NDVI time series.

\begin{table}[!ht]
\centering
\caption{Comparative analysis of MDA I and MDA II algorithms for grassland event detection using i) no interpolation, ii) Linear Interpolation (LI), iii) Akima Interpolation (AI) iv) Quadratic Interpolation (QI), and v) the SF model.}
\label{tab:mowing_detection_performance}
\begin{adjustbox}{width=\columnwidth,center}
\begin{tabular}{@{}p{0.2\columnwidth}>{\centering\arraybackslash}p{0.2\columnwidth}>
{\centering\arraybackslash}p{0.2\columnwidth}>
{\centering\arraybackslash}p{0.2\columnwidth}>
{\centering\arraybackslash}p{0.2\columnwidth}@{}}
\toprule
{Algorithm} & {Interpolation} & {Recall} & {Precision} & {F1-Score} \\ 
\midrule
\multirow{5}{*}{{MDA I}} & 
{-}  & 0.755  & 0.717 & 0.736    \\
& {LI} & 0.719  & 0.800 & 0.757 \\
& {AI} & 0.712  & 0.803 & 0.755 \\
& {QI} & 0.732 & 0.800 & 0.765 \\
& {SF} & 0.749  & 0.928 & 0.829 \\
\midrule
\multirow{5}{*}{{MDA II}} & 
{-} & 0.762  & 0.730 & 0.746  \\
& {LI} & 0.635 & 0.843 & 0.724 \\
& {AI} & 0.696 & 0.857 & 0.768 \\
& {QI} & 0.707 & 0.859 & 0.776 \\
& {SF} & 0.770 & 0.923 & 0.840  \\
\bottomrule

\end{tabular}
\end{adjustbox}
\end{table}

\begin{figure}[!ht]
\centering
\includegraphics[width=0.95\columnwidth]{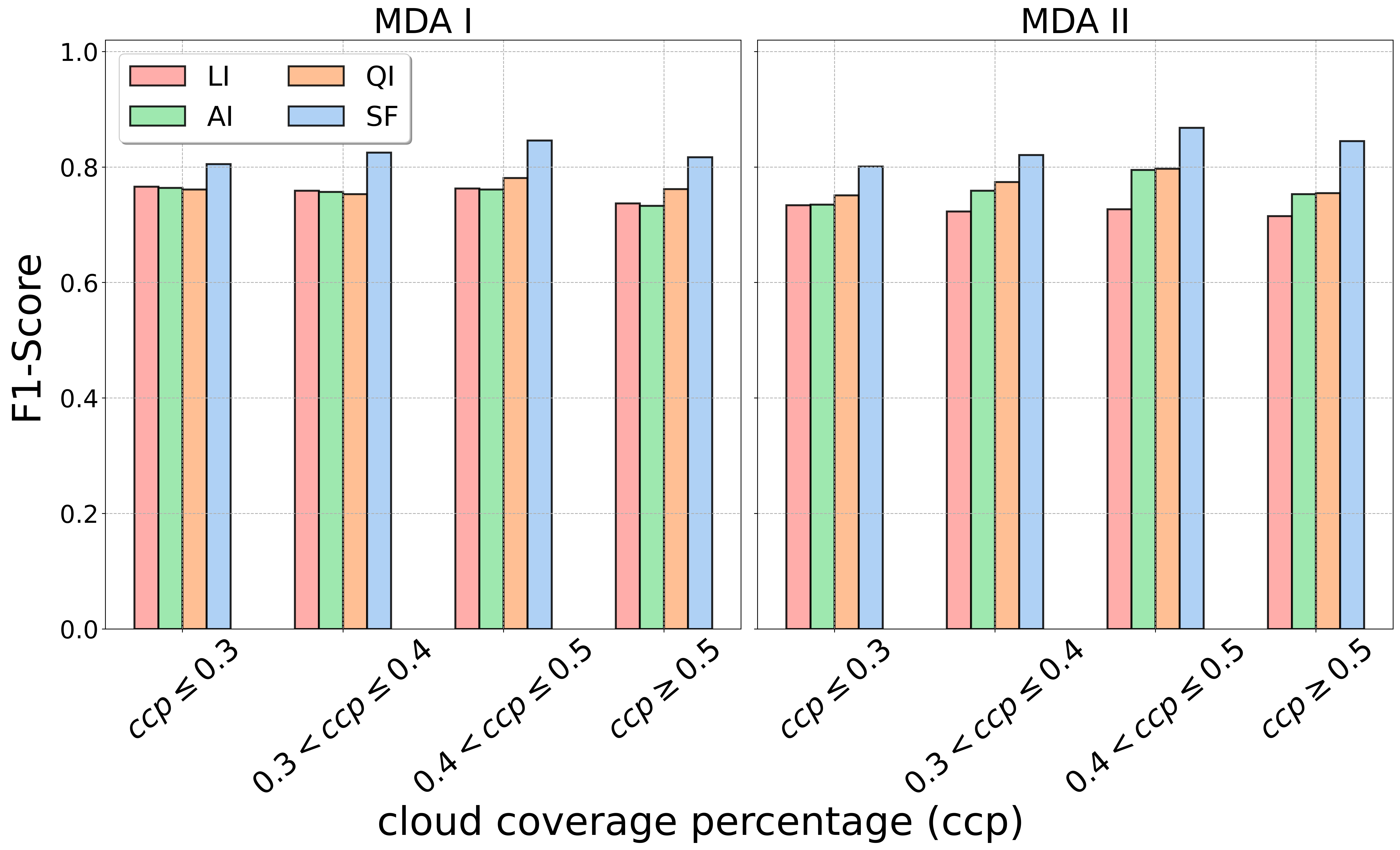}
\caption{F1-score for mowing event detection and for different cloud coverage percentage (ccp) scenarios. For further analysis, the results are detailed in Table S7 in \ref{sec:supplementary_data}.}
\label{fig:mowing_ccp}
\end{figure}

In Fig \ref{fig:mowing_ccp}, we present F1-score categorized by cloud coverage percentage (ccp). We divided the groups using the mean ($\mathrm{\mu} \simeq 0.4$) and standard deviation ($\mathrm{\sigma} \simeq 0.1$) values of the cloud coverage distribution, which resembled a normal distribution. Based on this statistical analysis, we constructed four distinct categories accordingly: ccp $\mathrm{< \mu - \sigma}$ (101 cases), $\mathrm{\mu - \sigma \leq}$ ccp $\mathrm{< \mu}$ (252 cases), $\mathrm{\mu \leq}$ ccp $\mathrm{< \mu + \sigma}$ (291 cases), ccp $\mathrm{\geq \mu + \sigma}$ (159 cases). The figure reveals that SF consistently delivered superior performance across various cloud coverage scenarios, highlighting its robustness against any cloud coverage situation. This is becoming more evident in cases with very high cloud coverage percentages. This can be attributed to SF's ability to handle cases resembling extended cloud coverage, which often obscure mowing events. The NDVI time series constructed based on SAR data excels in recovering such challenging cases, resulting in better performance.

\begin{figure}[!ht]
\centering
\includegraphics[width=0.95\columnwidth]{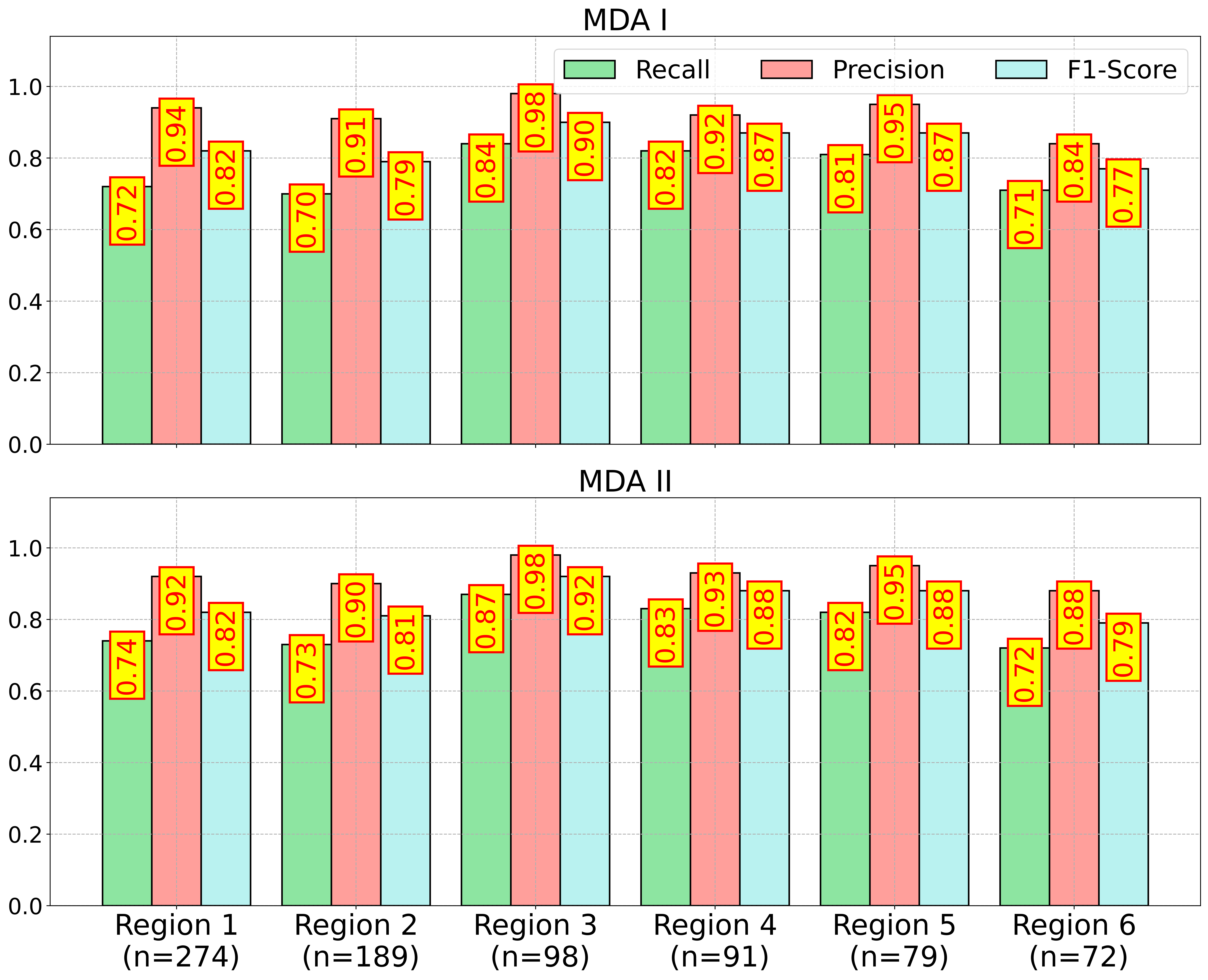}
\caption{Mowing detection performance (i.e., recall, precision and F1-score) over the different study regions for MDA I and MDA II, using NDVI time series produced by the proposed SF model.}
\label{fig:mowing_per_region}
\end{figure}

Fig. \ref{fig:mowing_per_region} illustrates the recall, precision, and F1-score obtained from both MDA I and MDA II across each of the study regions. Once again, the results obtained from MDA II exhibit a slight improvement over MDA I. Notably, Region 3 stands out with optimal results with precision values close to 1 and very high recall rates, which results in F1-score surpassing 0.9 for both MDAs. Similar behavior is exhibited in Regions 4 and 5, too. On the other hand, the rest of the Regions present worse performance, with an F1-score around 0.80. It is important to highlight that overall, precision can reach significantly high values, occasionally approaching 1, when we utilize the NDVI time series constructed with the SF model.
This effect can be attributed to the SF model's ability to correct cloudy observations and thus prevent the MDAs from incorrectly identifying these cases as mowing events.

\begin{figure}[!ht]
\centering
\includegraphics[width=0.9\columnwidth]{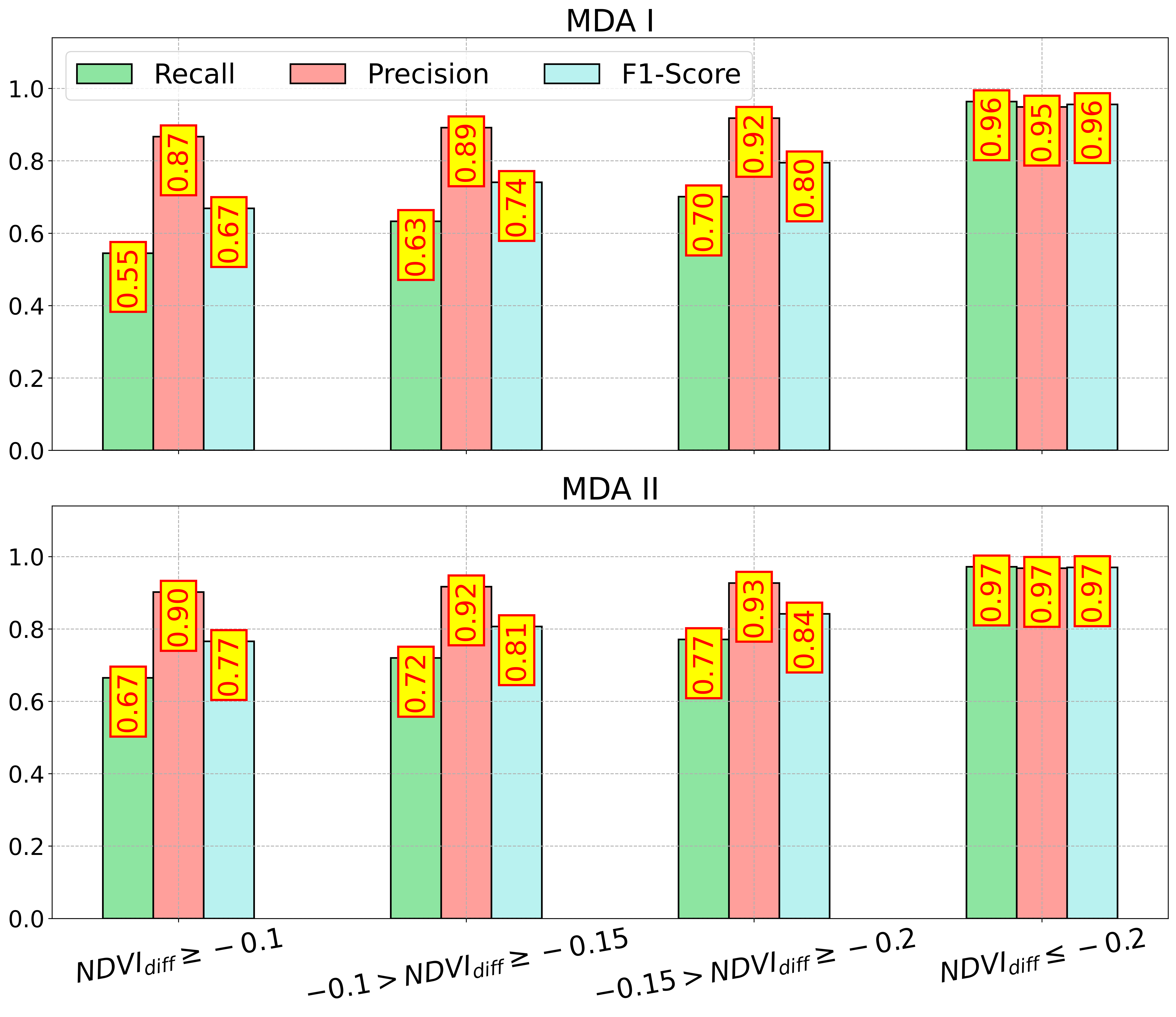}
\caption{Mowing detection performance (i.e., recall, precision and F1-score) across NDVI differences between successive timestamps for MDA I and MDA II, using NDVI time series produced by the proposed SF model.}
\label{fig:mowing_per_ndvi}
\end{figure}

Fig. \ref{fig:mowing_per_ndvi} depicts the recall, precision, and F1-score obtained from both MDA I and MDA II for NDVI differences between successive timestamps. Detection accuracy improves with larger NDVI differences, reaching near-perfect results for differences with absolute values greater than 0.2. However, for smaller drops (less than 0.1), detection becomes more challenging, particularly for MDA I, which identifies slightly more than half of the actual cases (recall of 55\%). This issue does not necessarily reflect the quality of the NDVI time series produced by the model, but rather the sensitivity of the mowing detection algorithms to minor NDVI changes. As anticipated, due to the design of the applied methodologies, their efficiency decreases with lower NDVI differences, underscoring the need for adaptive thresholds.

\begin{figure}[!ht]
\centering
\includegraphics[width=0.85\columnwidth]{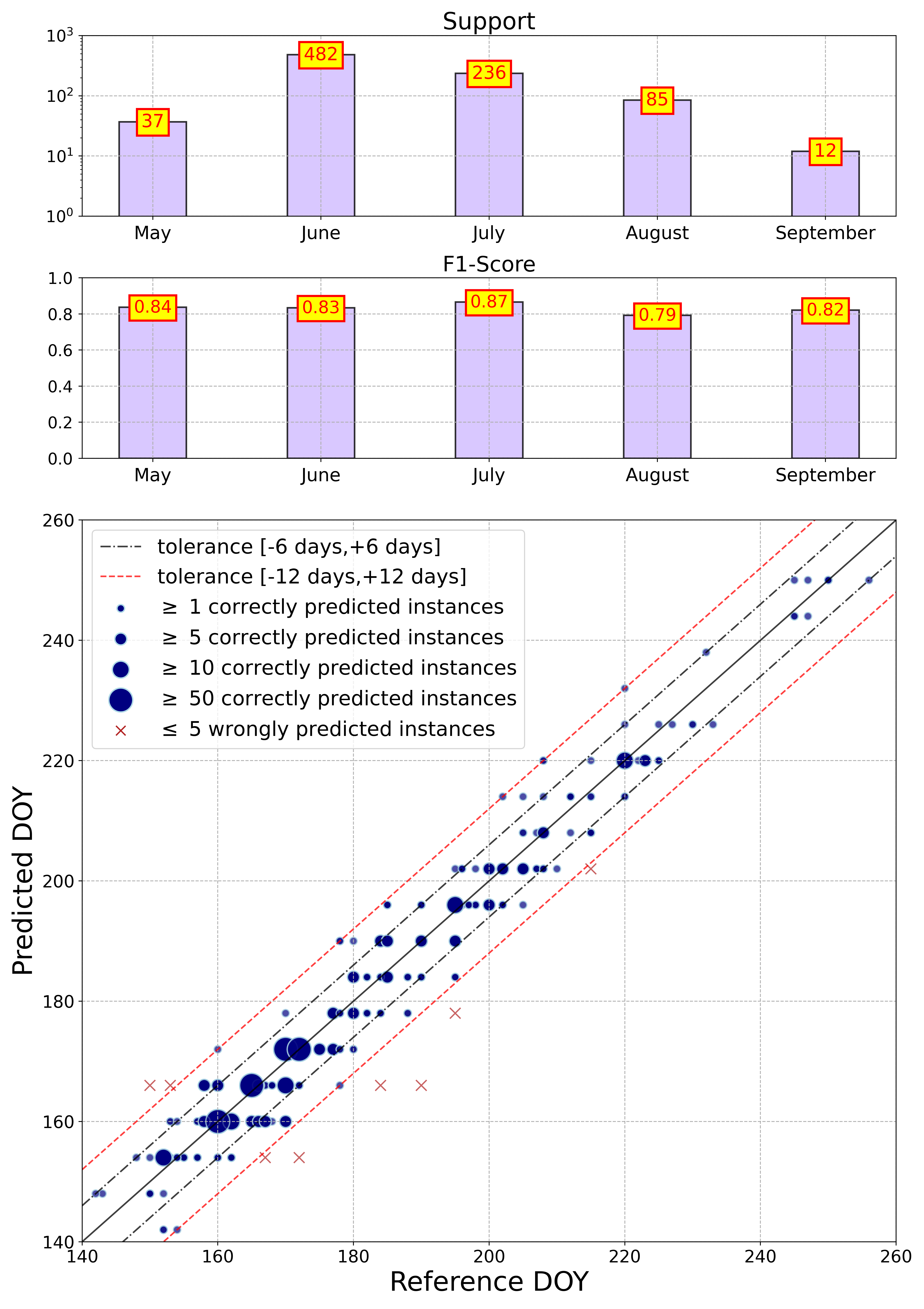}
\caption{Scatter plot comparing reference and predicted Day of Year (DoY) for grassland events using MDA II. Point size indicates the number of observations. The top bar chart shows the number of events and the bottom bar chart shows the F1-Score per month.}
\label{fig:scatter_schwieder}
\end{figure}

In Fig. \ref{fig:scatter_schwieder}, a scatter plot is presented that compares the ground truth and predicted values using MDA II, which has been proven to be slightly more accurate than MDA I. Only absolute differences less than 12 days are considered accurate, as larger values are deemed too significant or distant. The size of each point indicates the number of observations on this reference day. 
The plot indicates a linear relationship between the measures, with an $\mathrm{R^2}$ value of 0.946. Most events occur in June and July, as indicated by the top bar plot. Misclassifications mainly arise when the model detects events slightly later than their actual occurrence, possibly due to delayed NDVI decreases resulting from prior missing values.
The bottom bar chart presents the F1-scores grouped by month, revealing small fluctuations. This consistency in F1-scores across months highlights the model's robustness in event timing prediction.

\begin{figure}[!ht]
\centering
\includegraphics[width=\columnwidth]{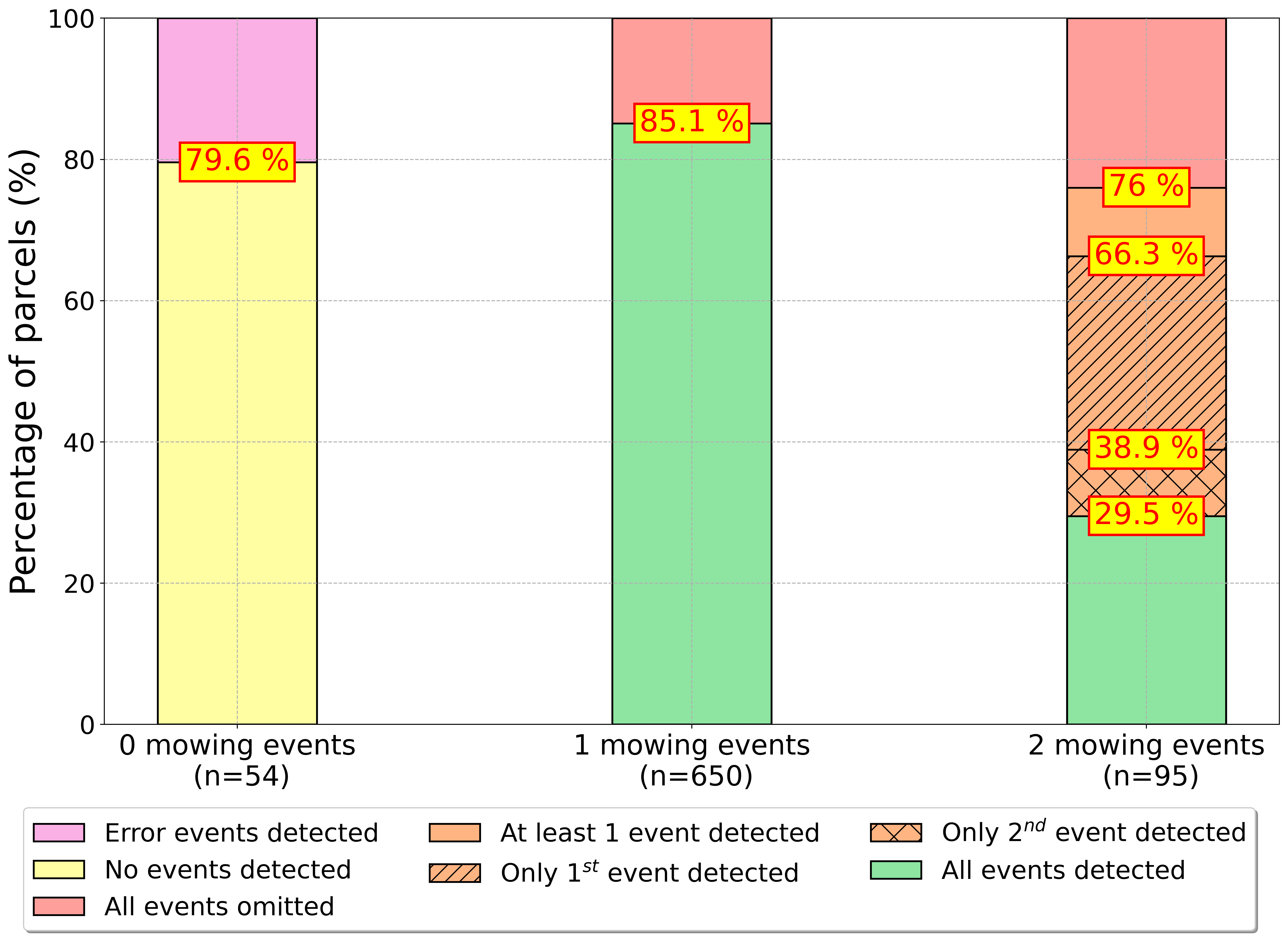}
\caption{Evaluation of the performance of MDA II on the different number of mowing events that occurred in parcels during the season. The first bar corresponds to the performance against non-mowed parcels, the second bar against parcels that were mowed exactly one time and the third bar for parcels that were mowed twice. For each of them, the percentage of actual cases correctly identified (i.e., recall) or the portion of the actual events omission is illustrated.}
\label{fig:schwieder_freq_analysis}
\end{figure}

Fig. \ref{fig:schwieder_freq_analysis} illustrates the recall for MDA II across parcels annotated with different numbers of total mowing events. Among the 54 unmown parcels, approximately 79.6\% were correctly identified as unmown. Furthermore, in parcels with a single mowing event, the recall percentage showed significantly improved performance, exceeding 85\%. However, for the 95 parcels with two mowing events, the algorithm accurately identified both events in only 29.5\% of cases. In particular, the recall rate for the first mowing event was 66\%, whereas for the second event, the recall rate dropped to 39\%.  This discrepancy could be explained if we combine Fig. \ref{fig:mowing_per_ndvi} and Fig. S9. Second mowing event cases presented much lower absolute values of NDVI drops (lower than 0.1), where the respective performance of MDA II is worse. Notably, in approximately 76\% of these two-event parcels, at least one event was detected accurately. Similar results for parcels with two mowing events have also been reported by \cite{devroey2022} across Wallonia.

\subsection{Recovery of masked events} \label{sec:masked_events}

In Section \ref{sec:4_2}, we assessed the impact of the SF algorithm on the two event detection methods. This section investigates cases where grassland parcel activities are unobservable through optical data (due to consecutive cloudy observations), rendering NDVI-based detection inadequate for identifying such events. For this purpose, we selected 359 cases from the dataset with a single mowing event in order to intentionally hide them. In these cases, we artificially masked out the NDVI value from the day before the event and the following values that could reveal grassland activity (e.g., lower NDVI values). This led to gaps of 3-7 consecutive missing values (24-48 days). 

In Table \ref{tab:mowing_detection_performance_hidden}, we compare the performance of each mowing detection algorithm applied to i) the raw NDVI series (no interpolation) and ii) the reconstructed NDVI time series using the SF model. Both algorithms performed poorly without interpolation due to a lack of optical information during the events. On the other hand, SF was able to recover nearly half of the actual events for MDA I and over half for MDA II, with relatively high F1-score values of 0.63 and 0.68, respectively. This demonstrates the reliability of SF in improving detection accuracy and reconstructing NDVI drops, even for large temporal gaps, making it a promising approach when optical data fails to capture event information. Additionally, in Fig. S10 and S11 (refer to section \ref{sec:supplementary_data}), spatial and temporal distribution for the recall is illustrated respectively, highlighting the robustness of the results.

\begin{table}[!ht]
\centering
\caption{Performance of the MDA I and MDA II algorithms on detecting artificially masked out grassland mowing events, using the actual NDVI values and the reconstructing ones by utilizing the SF algorithm.}
\label{tab:mowing_detection_performance_hidden}
\begin{adjustbox}{width=\columnwidth,center}
\begin{tabular}{@{}p{0.21\columnwidth}>{\centering\arraybackslash}p{0.21\columnwidth}>
{\centering\arraybackslash}p{0.15\columnwidth}>
{\centering\arraybackslash}p{0.15\columnwidth}>
{\centering\arraybackslash}p{0.15\columnwidth}@{}}
\toprule
 {Algorithm} & {Interpolation} & {Recall} & {Precision} & {F1-Score}\\ 
 \midrule
 \multirow{2}{*}{MDA I} & {-} & 0.036 & 0.074 & 0.048 \\
  & {SF} & 0.521 & 0.810 & 0.634 \\
 \midrule
 \multirow{2}{*}{MDA II} & {-} & 0.123 & 0.148 & 0.134 \\
  & {SF} & 0.655 & 0.701 & 0.677  \\
 \bottomrule
\end{tabular}
\end{adjustbox}
\end{table}

\begin{figure}[!ht]
\centering
\includegraphics[width=0.8\columnwidth]{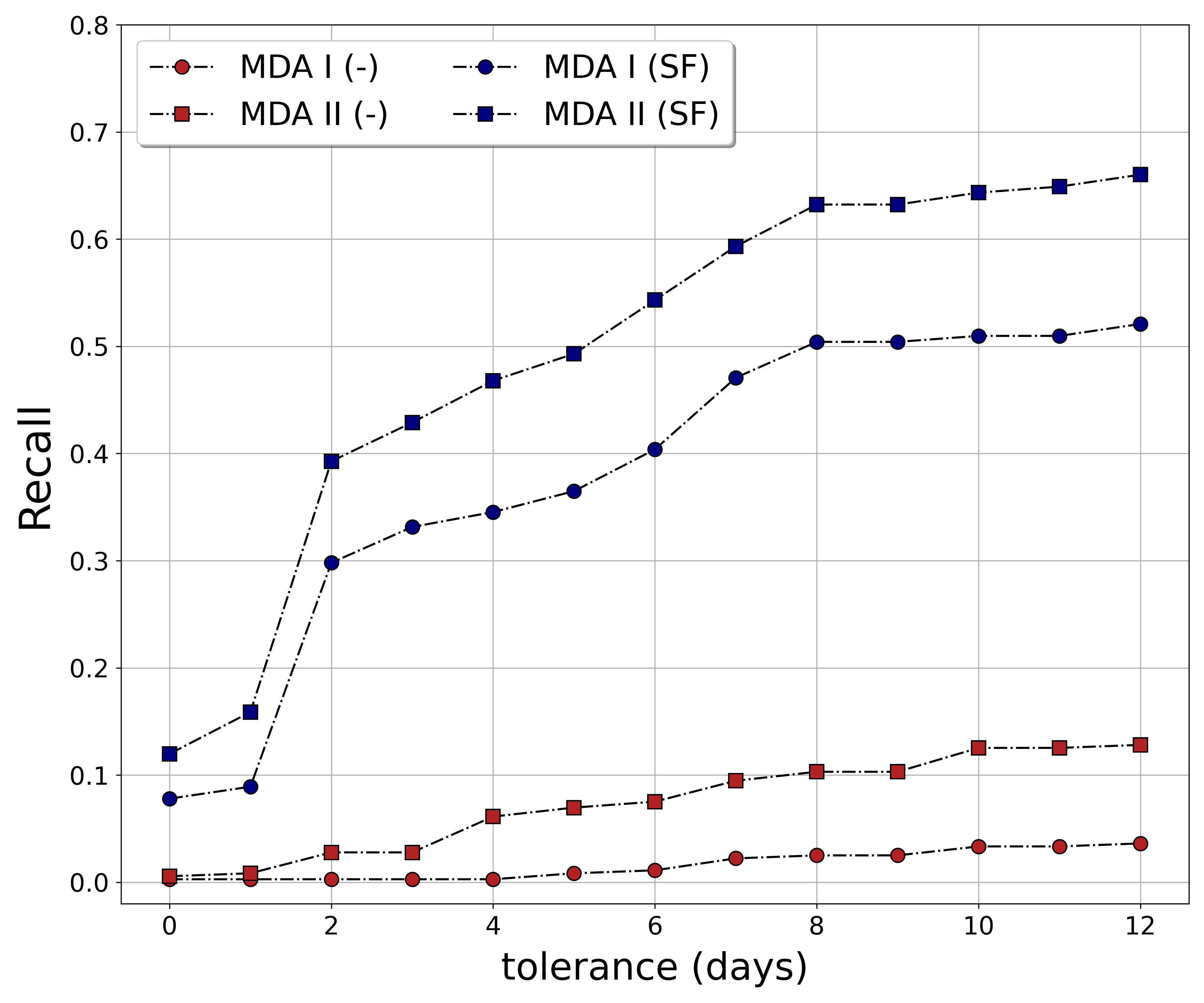}
\caption{Evolution of recall score using  MDA I and MDA II algorithms and different time-windows (tolerance). Tolerance refers to the difference in days between the actual and the predicted event for which we consider the prediction as correct. Blue dots/squares indicate the results of MDA I/MDA II using the SF algorithm for interpolation while the red dots/squares indicate the results of MDA I/MDA II without interpolation.}
\label{fig:f_score_hidden}
\end{figure}

In Fig. \ref{fig:f_score_hidden}, we illustrate how the accuracy of event detection is influenced by the time-window (tolerance) used to define an event as correct. The SF method reached its optimal recall performance of around 0.5 for MDA I and 0.6 for MDA II after 8 days, which remained relatively stable for higher tolerance parameters as we were reaching the default threshold (i.e., 12 days). Interestingly, even for very low tolerance values, it achieved satisfactory performance, especially considering cases of completely hidden events. In contrast, non-interpolated NDVI time series resulted in extremely low recall values, irrespective of the tolerance value.

\begin{figure}[!ht]
\centering
\includegraphics[width=0.8\columnwidth]{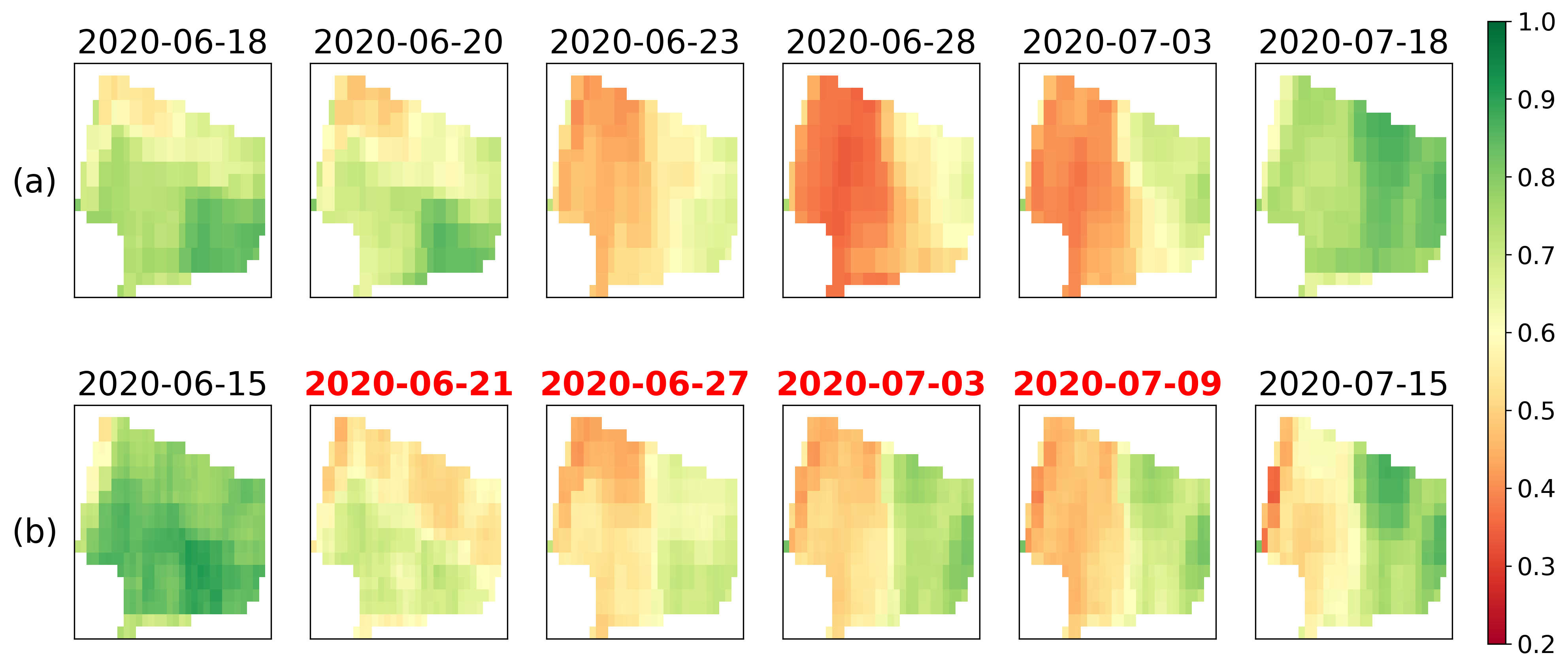}
\caption{Spatial visualization of NDVI time series for a parcel. (a) Actual NDVI extracted from the Sentinel-2 cloud-free acquisition dates, (b) artificial NDVI time series created after masking the values on the dates during and after the observed mowing event, which is highlighted in red (from 21/06 to 09/07)). For further visualization, refer to Fig. S12 of \ref{sec:supplementary_data}.}
\label{fig:mowing_example_2}
\end{figure}

Fig. \ref{fig:mowing_example_2} provides a visual demonstration of the effectiveness of the SF algorithm in detecting mowing events in grassland areas. Specifically, it displays the actual NDVI values of a parcel during the event dates, as well as the predicted values generated by the SF algorithm. We observe a high correlation between the predictions and the actual measurements, indicating the model's effectiveness in detecting mowing events. These findings are significant for grassland sustainability as they offer detailed information about the extent and location of mowing events.

\subsection{Deep learning mowing detection algorithm}

In previous sections, we have demonstrated the effectiveness of the SF algorithm in accurately reconstructing missing NDVI values in grasslands. The algorithm performed well even when missing values occurred during mowing events. We have also employed two threshold-based algorithms to predict mowing events and evaluated the impact of using the SF algorithm on the input NDVI time series to improve prediction performance.

In this section, we utilized the SF architecture to detect mowing events (see Section \ref{sec:event_detection_methods}) and compared its performance with the algorithms mentioned earlier. To evaluate the model, we conducted a 3-fold cross-validation using the photo-interpretation dataset (refer to Section \ref{sec:photointerpretation}). 
The dataset was divided into three folds at the parcel level, taking into account the date of the annotated mowing events to guarantee fairness. Specifically, each fold contained cases with event date distributions that mirrored those of the entire dataset. However, even though we split on the parcel level, the training of the models was performed at the pixel level.
This approach helped us obtain reliable results for evaluating the SF architecture. 
Finally, for our proposed approach, which required a 6-day interval time series, we trained two models: i) one using time series interpolated with the Akima method, and ii) another using time series constructed using the SF model. These results were compared with those obtained from each MDA under the following conditions: i) using raw, unprocessed time series without interpolation, ii) using the same time series after Akima interpolation to achieve a 6-day temporal resolution, and iii) using the corresponding results after applying the SF model.

Table  \ref{tab:mowing_detection_cnn_rnn} displays the results of the mowing event detection task. 
MDA I achieved the highest recall (0.875), albeit with relatively low precision (0.686), resulting in suboptimal overall performance. Notably, Akima interpolation significantly improved precision without sacrificing recall. A similar trend was observed for MDA II, with a slight decrease in overall recall, leading to a substantial increase in precision. In both cases, the use of SF-derived time series resulted in significantly improved precision while maintaining similar recall rates, leading to an increased overall F1-score of approximately 0.83.

\begin{table}[!ht]
\centering
\caption{Recall, precision, and F1-score of MDA I, MDA II, and SF algorithm for the detection of grassland mowing events. The values correspond to the mean and the standard deviation scores by using a 3-fold cross-validation.}
\label{tab:mowing_detection_cnn_rnn}
\begin{adjustbox}{width=\columnwidth,center}
\begin{tabular}{@{}p{0.19\columnwidth}>{\centering\arraybackslash}p{0.18\columnwidth}>
{\centering\arraybackslash}p{0.24\columnwidth}>
{\centering\arraybackslash}p{0.24\columnwidth}>
{\centering\arraybackslash}p{0.24\columnwidth}@{}}
\toprule
{Algorithm} & {Interpolation} &  {Recall} & {Precision} & {F1-Score} \\ 
\midrule
{MDA I} & {-} & 0.875 $\pm$ 0.003 & 0.686 $\pm$ 0.010 & 0.769 $\pm$ 0.007 \\
{MDA I} & {AI} & 0.841 $\pm$ 0.005 & 0.759 $\pm$ 0.012 & 0.798 $\pm$ 0.009 \\
{MDA I} & {SF} & 0.821 $\pm$ 0.005 & 0.834 $\pm$ 0.005 & 0.827 $\pm$ 0.004 \\
\midrule
{MDA II} & {-} & 0.817 $\pm$ 0.006 & 0.763 $\pm$ 0.015 & 0.789 $\pm$ 0.009 \\
{MDA II} & {AI} & 0.811 $\pm$ 0.008 & 0.803 $\pm$ 0.014 & 0.807 $\pm$ 0.010 \\
{MDA II} & {SF} & 0.814 $\pm$ 0.007 & 0.849 $\pm$ 0.011 & 0.831 $\pm$ 0.006 \\
\midrule
{Ours'} & {AI} & 0.829 $\pm$ 0.009 & 0.830 $\pm$ 0.009 & 0.829 $\pm$ 0.001 \\
{Ours'} & {SF} & 0.840 $\pm$ 0.004 & 0.855 $\pm$ 0.009 & 0.847 $\pm$ 0.004 \\
\bottomrule
\end{tabular}
\end{adjustbox}
\end{table}

Our suggested approach outperformed threshold-based methods in identifying mowing events, especially with SF-processed NDVI input, achieving optimal results with significantly higher precision (0.83 and 0.86). In both cases of our proposed implementation, the mean F1-score exceeded 80\%, indicating highly effective performance in mowing event detection. Remarkably, our suggested supervised approach achieved F1-score close to 85\% when incorporating SF for interpolation, which marks an improvement of approximately 2\% over the Akima case, highlighting the added value of NDVI reconstruction.
This represents a notable improvement, considering the utilization of highly dense NDVI time series (i.e., minimal cloud coverage), making the detection of the majority of events relatively easy for both input scenarios.

\section{Discussion}\label{sec:discussion}

\subsection{Cloud filtering for event detection}

Cloud cover can be so extensive at times that it results in gaps in the time series that may span more than a month. In Section \ref{sec:masked_events}, we illustrated how the SF algorithm produces a continuous time series, allowing for the identification of mowing events that would have been missed if only optical imagery had been used. However, cloud masks may not always detect the presence of clouds accurately, resulting in false measurements and inaccurate event detection. According to \cite{hardy2021sen2grass}, there is a trade-off between data availability and accuracy when setting the cloud cover threshold for grassland monitoring systems, which depends on the end-users preferences.

Apart from reconstructing missing information, the SF methodology also serves as a supplementary cloud filtering mechanism. By incorporating SAR and optical data, we observe that it can identify and disregard available Sentinel-2 measurements that result in sudden drops of NDVI when no correlation between the two components is justified. 

\begin{figure}[!ht]
\centering
\includegraphics[width=0.8\columnwidth]{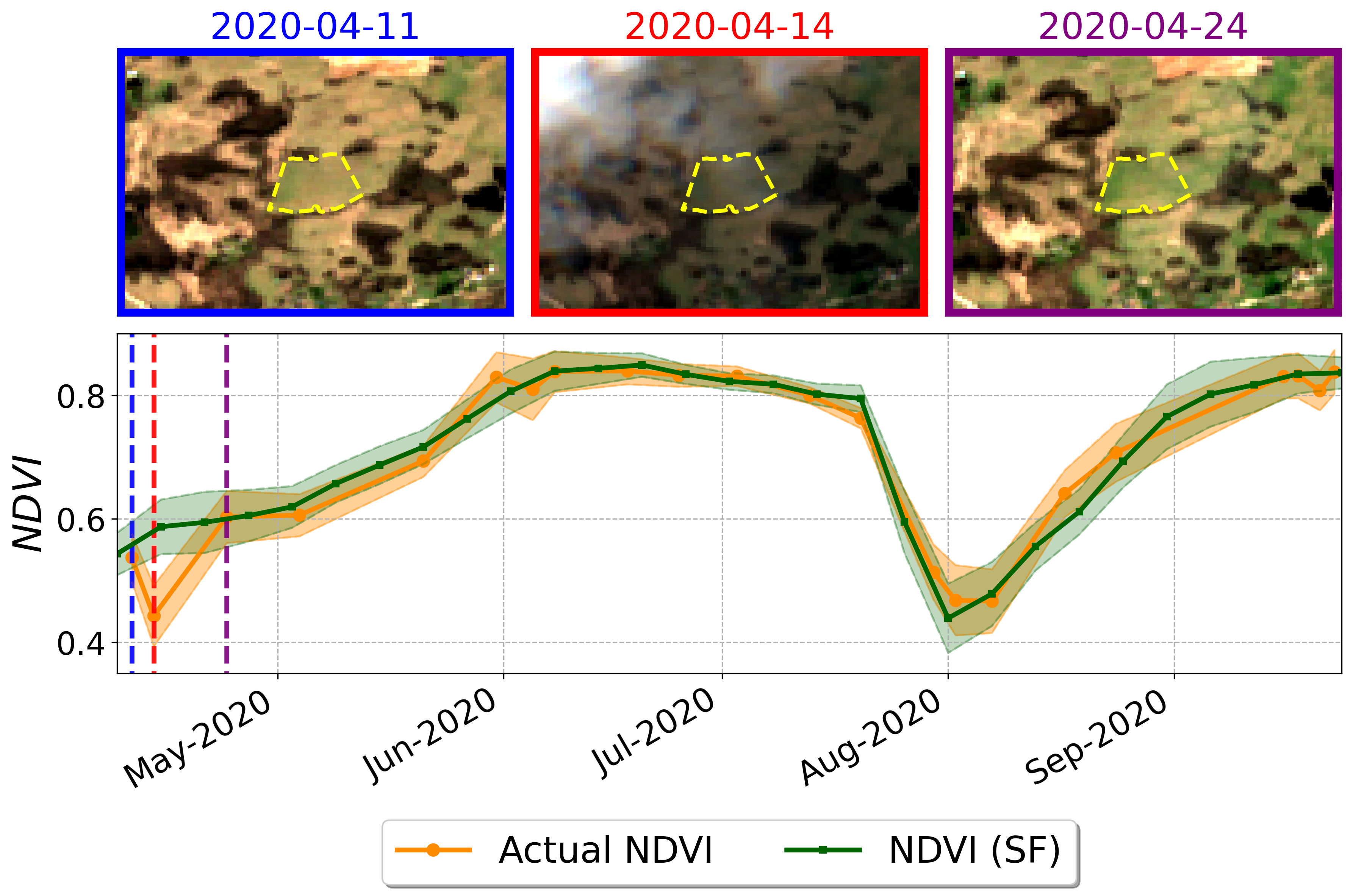}
\caption{Example of a case where the SF algorithm acts as a cloud filtering mechanism. The top three RGB images have been derived from consecutive Sentinel-2 captures. The first and the last are clear-sky images while the middle one suffers from high cirrus clouds. In the plot below are presented the actual NDVI time series (orange) and the predicted ones using the SF algorithm (green). The blue, red, and purple vertical dotted lines refer to the dates that the aforementioned images were captured.}
\label{fig:filter_drops}
\end{figure}

In Fig \ref{fig:filter_drops}, we illustrate a typical scenario where the sen2Cor algorithm fails to detect high cirrus clouds (mid-April), resulting in the inclusion of a noisy measurement that could potentially be misinterpreted as an event from the mowing event algorithms.
As shown, the SF methodology ignores the NDVI drop, even though it was included in the algorithm's input data and instead, it generates an output following the general upward trend.

Taking the aforementioned into account, a filtering routine can be designed. By employing a particular threshold difference between the actual NDVI value and the one generated by the SF model, timestamps that are potentially affected by clouds can be identified and excluded. This additional technique can improve the initial cloud masking process, resulting in more accurate time series data for monitoring agricultural land and making it more suitable for subsequent tasks.

\subsection{Gap-filling ablation study} \label{sec:ablation study}

To determine the individual contributions of different components within the SF model, we conducted an ablation study. This study specifically analyzed the NDVI gap-filling performance across seven varied input data configurations. 
The baseline scenario includes all the features described earlier in Table \ref{tab:features}, while the other six scenarios are constructed by combining specific features, and their derived products. These scenarios involve combinations or single deployments of the $\mathrm{\sigma_0}$ (similar to \cite{zhao2020}), coherence, and NDVI features to showcase various input configurations. All scenarios maintained consistent model parameter configurations and used the same training and validation pixels from the datasets (see Table \ref{tab:pixels_distro}). Table \ref{tab:ablation study} presents the MAE and $\mathrm{R^2}$ of the different scenarios.
The performance of the model without utilizing the cloud-free NDVI time series as input is substantially lower. In these cases, the average MAE exceeds 0.051, and the $\mathrm{R^2}$ remains below 0.79. Incorporating SAR data with the available NDVI measurements led to a substantial improvement, reducing the average MAE by 26-29\% depending on the scenario, and increasing $\mathrm{R^2}$ by approximately 0.09, corresponding to an enhancement of about 11\%. Specifically, MAE gets average values below 0.04, and $\mathrm{R^2}$ over 0.87. Notably, even when using only the NDVI component, there was a significant enhancement in terms of MAE, with a decrease of 0.01 compared to using only SAR input features. This outcome could be justified, as the model learns from the statistical profile of regional NDVI, operating as a deep learning-oriented interpolation approach, unaffected by less-correlated SAR data. Consequently, the results are similar to or outperform traditional baseline interpolation methodologies, achieving the best MAE of 0.43 (see Section \ref{sec:results}). Finally, the best results were obtained with the inclusion of all derived features, pointing out the contribution of SAR components to the model.

\begin{table}[!ht]
\centering
\caption{MAE and $\mathrm{R^2}$ obtained from the SF model on the masked timestamps using different combinations of the input features.}
\label{tab:ablation study}
\begin{adjustbox}{width=\columnwidth,center}
\begin{tabular}{@{}p{0.38\columnwidth}>{\centering\arraybackslash}p{0.35\columnwidth}>{\centering\arraybackslash}p{0.35\columnwidth}@{}}
\toprule
 {Input Features} & {MAE} & {$\mathrm{R^2}$}\\ 
\midrule
\{$\mathrm{coherence}$\} & 0.0533 $\pm$ 0.007 & 0.785\\
\{$\mathrm{\sigma_0}$\} & 0.0529 $\pm$ 0.006 & 0.787\\
\{$\mathrm{coherence, \sigma_0}$\} & 0.0518 $\pm$ 0.006 & 0.789\\ 
\{$\mathrm{NDVI}$\} & 0.0413 $\pm$ 0.008 & 0.841\\
\{$\mathrm{NDVI, coherence}$\} & 0.0395 $\pm$ 0.004 & 0.876\\
\{$\mathrm{NDVI, \sigma_0}$\} & 0.0389 $\pm$ 0.004 & 0.876\\
\{$\mathrm{NDVI, coherence, \sigma_0}$\} & 0.0364 $\pm$ 0.004 & 0.885\\
\bottomrule
\end{tabular}
\end{adjustbox}
\end{table}

Additionally, to thoroughly understand the role of different elements in our CNN-RNN architecture, we conducted an ablation study by systematically removing or modifying parts of the network and evaluating the performance impact. This approach allowed us to determine the optimal hyperparameter setup. In Table S4 (see \ref{sec:supplementary_data}), detailed results of the tested architecture variants are presented in terms of MAE and $\mathrm{R^2}$. Specifically, the contribution of the CNN layers at the beginning of the network is evident, as they significantly enhanced the overall performance. The number of parameters for each model configuration is also provided, emphasizing that our models are lightweight and suitable for deployment on conventional infrastructures, with no requirement for GPU acceleration.

Finally, in Table S5 (see \ref{sec:supplementary_data}), we present the results of a grid search evaluating different training weight parameters ($\mathrm{w_{\alpha}}$, $\mathrm{w_{\beta}}$) for both masked and non-masked input NDVI values. The results indicate that smaller $\mathrm{w_{\beta}}$ values compared to $\mathrm{w_{\alpha}}$, led to optimal performance. Conversely, assigning equal weight parameters (i.e., setting both $\mathrm{w_{\alpha}}$ and $\mathrm{w_{\beta}}$ to 1) resulted in slightly poorer MAE, while worse results are obtained with larger $\mathrm{w_{\beta}}$ parameters. These findings underscore our initial statement to prioritize attention on masked timestamps to mitigate the NDVI leakage issue when used as both input and target values (see Section \ref{sec:model_training}).

\subsection{Spatial generalization} \label{sec:region_generalization}

To gain additional insights, we conducted a regional cross-validation experiment to assess the model's behavior when trained on a limited number of regions. This allowed us to evaluate the model's ability to generalize spatially and transfer to regions that were not included in the original training process. Additionally, by examining how the number of training samples affects the quality of the training process, we can determine the minimum number required to achieve satisfactory results.

\begin{figure}[!ht]
\centering
\includegraphics[width=0.85\columnwidth]{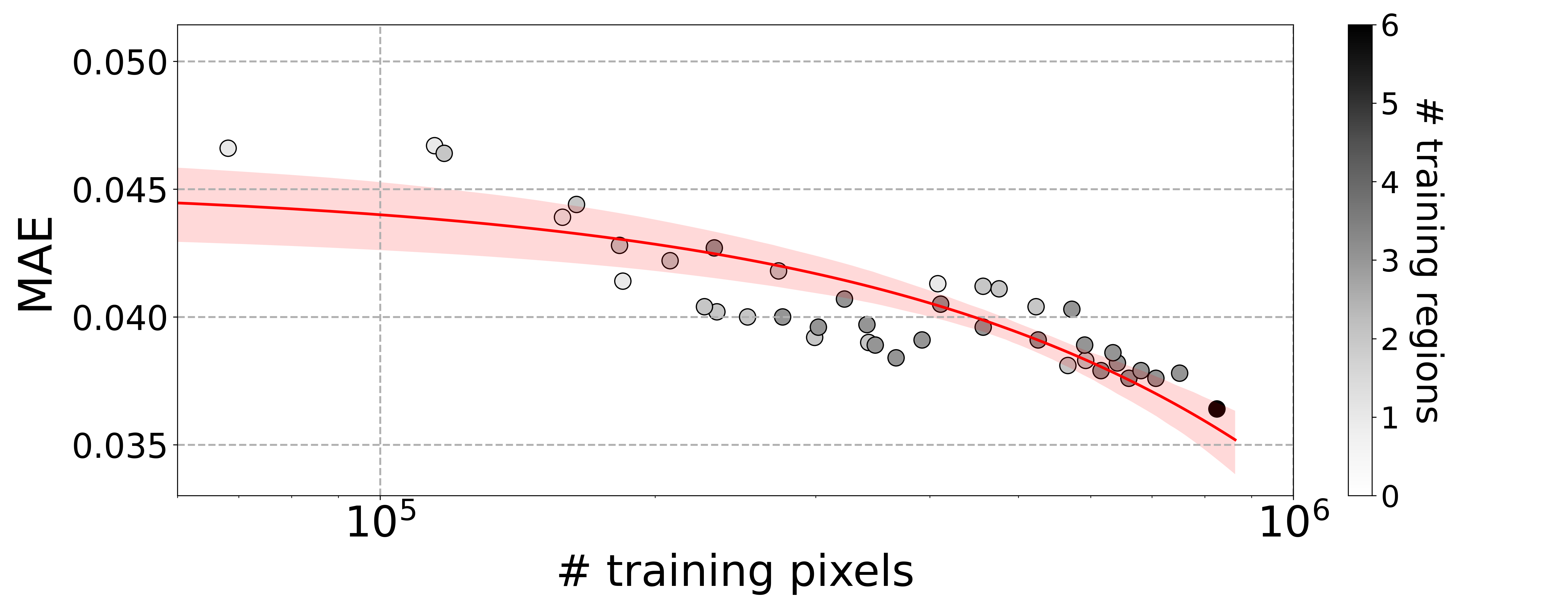}
\caption{Scatter plot comparing the MAE with different numbers of training pixels. The color of each point indicates the number of training regions, as shown in the right color bar.}
\label{fig:mae_training_size}
\end{figure}

The optimal generalization capability of the SF model depends heavily on the size of the training dataset. To accurately learn from the distribution of missing values and fill NDVI gaps, it is crucial to have a representative sample of training instances across the entire Lithuania. Fig. \ref{fig:mae_training_size} demonstrates how increasing the number of training instances and the regions used for training affects the MAE. 
The darker colors indicate more training regions used. While including more training samples led to a decrease in MAE, it appears that using training samples from different regions is more crucial. Interestingly, by utilizing only three regions, we can achieve near-optimal results. 

Despite the limited number of training regions, in many cases, the model achieved near-optimal performance. This may be due to the uniform mowing policy and similar grassland management practices throughout Lithuania, resulting in similar grassland vegetation profiles. It can be also attributed to the fact that some regions are larger (e.g., Region 1 and 2), which include diverse cases, and thus when they are used for training, we can more accurately represent the overall distribution. However, it's essential to consider the cloud coverage distribution of each training region, as this can affect the model's transferability beyond those regions. If the cloud coverage profile of the training samples differs significantly from the overall distribution, the model may overfit to the training regions and fail to generalize well. For more information, in the \ref{sec:supplementary_data} (Table S6) we present the MAE computed on the validation subset of each region, and overall, when the model was trained on one, two, or three regions.

Furthermore, to address concerns regarding the generalization capability of our model for the downstream task of mowing detection, we conducted an additional analysis of mowing detection performance across various MDA using the reconstructed NDVI produced from the trained models mentioned above. This analysis aimed to evaluate performance under different training region configurations, providing insights into the model's ability to detect events across diverse geographical regions. Our results corroborate this argument, demonstrating lower average F1-scores and increased variability when the SF model is trained using data from only one region. As the number of training regions increases, performance improves, approaching F1-scores comparable to those achieved after optimal SF model implementation. Specifically, for MDA I and MDA II, the average F1-scores reached 0.82 and 0.83, respectively, with three training regions included. 
For detailed information, refer to Table S8 in \ref{sec:supplementary_data}.

\subsection{Limitations and future work}

In recent years, numerous studies have concentrated on employing neural networks to combine SAR and optical data to tackle the problem of discontinuous optical image time series. However, none of these studies have investigated incorporating the available optical data directly into the model's architecture. According to \cite{garioud2021}, such an approach could potentially cause the model to ignore the remaining input features because the target and input data would be identical at several timestamps. Our observations during the training of SF using NDVI time series as an additional input feature confirmed this issue, as we noticed a high correlation between the input and output NDVI time series. To address this concern, we used temporal weighting for each input sample, as mentioned in Section \ref{sec:model_training}. We assigned masked NDVI values a weight of 0.75, while non-masked values received a weight of 0.25. Moreover, we assigned a weight of 0 to any value corresponding to an interpolated timestamp, as we used Akima interpolation to produce a complete time series. 
Additionally, an ablation study to determine the individual contributions of different components to the model showcases the clear benefits of including both optical and SAR information.

Our selection of the CNN-RNN architecture over more advanced deep learning architectures (e.g., Transformers) is motivated by the fact that it has been already suggested and applied in a similar research task \citep{zhao2020} and it also offers a computationally efficient approach. Furthermore, our primary objective was to emphasize the contribution of integrating the Sentinel-2 measurements into the model's feature space. Overall, we are confident that this architecture adequately presents our goal of improving the reconstruction of missing NDVI values and enhancing mowing detection accuracy. We also intend to explore a broader range of architectures in future research.

Utilizing neural networks' ability to learn complex temporal patterns, we extracted grassland phenology information and dependencies. Typically, NDVI requires approximately 10 days to recover half of the value lost after a grassland mowing event \citep{kolecka2018}.
In Lithuania, it is uncommon to find cases where no optical cloud-free images have been captured during such long periods, especially during the summer months.
The majority of these gaps occur before June, whereas most mowing events take place from early June onwards. Fig. S13 in \ref{sec:supplementary_data} depicts the temporal distribution of extended gaps. Extended gaps refer to occasions where pixels have been masked out because of cloud coverage for at least the three previous Sentinel-2 acquisitions. This duration is considered sufficient to conceal a mowing event. Consequently, while the model can effectively uncover hidden mowing events, it is important to note that such occurrences are anticipated to be relatively infrequent (5-10\%). This is also evident by combing Fig. S6 and S7 on \ref{sec:supplementary_data}.

Another challenge of the study relates to the photo-interpretation process employed for annotating mowing events. Despite efforts to ensure accuracy, human error remains inherent in manual interpretation tasks, potentially leading to inconsistencies or inaccuracies in event identification. Additionally, distinguishing between mowing and grazing events solely based on satellite imagery proved challenging. As a result, some events annotated as mowing may actually correspond to grazing activity, introducing a degree of ambiguity into the dataset. 

In Section \ref{sec:region_generalization}, we showcased the generalization ability of our model across Lithuania. Our future goal is to extend this approach to other European countries with diverse grassland mowing and activity policies, as well as varying cloud coverage distributions. This would lead to the creation of a pan-European model that could be highly valuable for CAP monitoring.

Moreover, our approach can be adapted through minor modifications to reconstruct other remote sensing indices, such as the Soil Adjusted Vegetation Index (SAVI), Normalized Burn Ratio (NBR), or even raw optical spectral bands. These indices are essentially beneficial for monitoring vegetation or detecting farming practices, including tillage and stubble burning. Consequently, a more versatile methodology can be developed to generate one or multiple relevant spectral indices, depending on the task. Furthermore, monitoring various land cover vegetation types is possible, including forests, arable croplands (such as wheat, maize, or barley), and orchards. 

Finally, the implementation of the SF algorithm (see Section \ref{sec:event_detection_methods}) for the detection of mowing events appears very promising, and we plan to conduct further evaluation of its potential in the future. Specifically, it will be additionally incorporated and tested within the “mowing detection inter-comparison exercise” (MODCiX). This initiative aims to evaluate various mowing detection algorithms using openly accessible satellite data across Europe under a common validation system.

\section{Conclusion}\label{sec:conclusion}

The Sentinel-2 satellite missions provide frequent optical images for the continuous monitoring of grassland activity. However, relying on optical images alone is often restricted by cloud coverage. To overcome this limitation, we proposed a CNN-RNN deep learning architecture that makes use of all available cloud-free optical images (Sentinel-2) and weather-independent SAR data (Sentinel-1), to produce continuous NDVI time series for grassland fields, with a temporal resolution of 6 days.
The integration of cloud-free Sentinel-2 data into our model is a significant advancement of this work, enhancing monitoring capabilities compared to relying solely on SAR data. The model has been trained on six different regions of Lithuania to account for different cloud coverage profiles and agro-climatic conditions.

Our method outperformed alternative gap-filling techniques (linear, Akima, and quadratic interpolation) in various evaluation scenarios,  demonstrating its transferability and generalization capacity. 
Most importantly, the generated continuous NDVI time series significantly enhanced the accuracy of mowing event detection by i) identifying events that would have otherwise gone undetected and ii) eliminating instances of false event predictions caused by clouds.
Our model demonstrated a promising F1-score of 84\%, surpassing the respective rates of baseline methodologies by more than 6\%.

\section*{CRediT authorship contribution statement}\label{sec:credit}

\textbf{Iason Tsardanidis:} 
Conceptualization, Methodology, Software, Validation, Formal Analysis, Investigation, Data Curation, Visualization, Writing – original draft, Writing – review \& editing. 
\textbf{Alkiviadis Koukos:} 
Conceptualization, Methodology, Validation, Formal Analysis, Data Curation, Visualization, Writing – review \& editing. 
\textbf{Vasileios Sitokonstantinou:}
Supervision, Conceptualization, Formal Analysis, Writing – review \& editing. 
\textbf{Thanassis Drivas:} 
Software, Data Curation, Writing – review \& editing. 
\textbf{Charalampos Kontoes:} 
Supervision, Writing – review \& editing, Funding acquisition.

\section*{Declaration of generative AI and AI-assisted technologies in the writing process}\label{sec:AI_writing}

During the preparation of this work, the authors used ChatGPT in order to improve readability. After using this tool, the authors reviewed and edited the content as needed and took full responsibility for the content of the publication.

\section*{Declaration of Competing Interest}\label{sec:competing_interest}

The authors declare that they have no known competing financial interests or personal relationships that could have appeared to influence the work reported in this paper.

\section*{Acknowledgements}\label{sec:acknowledgements}

This work has been supported by the ENVISION and CALLISTO projects, funded by European Union’s Horizon 2020 research and innovation programmes under grant agreements No. 869366 and No. 101004152.

The work of V. Sitokonstantinou has been partly supported by GVA PROMETEO AI4CS project on ‘AI for complex systems’ (2022-2026) -
CIPROM/2021/056.



\section*{Availability of data and Materials}\label{sec:data_availability}

Data is available at \href{https://zenodo.org/record/11651601}{https://zenodo.org/record/11651601}, a part of the code is accessible at \href{https://github.com/Agri-Hub/Deep-Learning-for-Cloud-Gap-Filling-on-Normalized-Difference-Vegetation-Index}{https://github.com/Agri-Hub/Deep-Learning-for-Cloud-Gap-Filling-on-Normalized-Difference-Vegetation-Index}.



\appendix

\section{Supplementary Material}
\label{sec:supplementary_data}

\bibliographystyle{elsarticle-num}
\bibliography{ref}

\begin{thebibliography}{10}
\expandafter\ifx\csname url\endcsname\relax
  \def\url#1{\texttt{#1}}\fi
\expandafter\ifx\csname urlprefix\endcsname\relax\def\urlprefix{URL }\fi
\expandafter\ifx\csname href\endcsname\relax
  \def\href#1#2{#2} \def\path#1{#1}\fi

\bibitem{mara2012}
F.~P. O'Mara, {The role of grasslands in food security and climate change}, Annals of Botany 110~(6) (2012) 1263--1270.
\newblock \href {https://doi.org/https://doi.org/10.1093/aob/mcs209} {\path{doi:https://doi.org/10.1093/aob/mcs209}}.

\bibitem{zhao2020grassland}
Y.~Zhao, Z.~Liu, J.~Wu, Grassland ecosystem services: a systematic review of research advances and future directions, Landscape Ecology 35 (2020) 793--814.
\newblock \href {https://doi.org/https://doi.org/10.1007/s10980-020-00980-3} {\path{doi:https://doi.org/10.1007/s10980-020-00980-3}}.

\bibitem{d'Adrimont2018}
R.~D’Andrimont, G.~Lemoine, M.~Van~der Velde, Targeted grassland monitoring at parcel level using sentinels, street-level images and field observations, Remote Sensing 10 (2018) 1300.
\newblock \href {https://doi.org/https://doi.org/10.3390/rs10081300} {\path{doi:https://doi.org/10.3390/rs10081300}}.

\bibitem{reinermann2023}
S.~Reinermann, S.~Asam, U.~Gessner, T.~Ullmann, C.~Kuenzer, Multi-annual grassland mowing dynamics in germany: spatio-temporal patterns and the influence of climate, topographic and socio-political conditions, Frontiers in Environmental Science 11 (2023).
\newblock \href {https://doi.org/https://doi.org/10.3389/fenvs.2023.1040551} {\path{doi:https://doi.org/10.3389/fenvs.2023.1040551}}.

\bibitem{ali2016}
I.~Ali, F.~Cawkwell, E.~Dwyer, B.~Barrett, S.~Green, {Satellite remote sensing of grasslands: from observation to management}, Journal of Plant Ecology 9~(6) (2016) 649--671.
\newblock \href {https://doi.org/https://doi.org/10.1093/jpe/rtw005} {\path{doi:https://doi.org/10.1093/jpe/rtw005}}.

\bibitem{Reinermann2020}
S.~Reinermann, S.~Asam, C.~Kuenzer, Remote sensing of grassland production and management—a review, Remote Sensing 12~(12) (2020) 1949.
\newblock \href {https://doi.org/https://doi.org/10.3390/rs12121949} {\path{doi:https://doi.org/10.3390/rs12121949}}.

\bibitem{gomez2017}
M.~{Gómez Giménez}, R.~{de Jong}, R.~{Della Peruta}, A.~Keller, M.~E. Schaepman, Determination of grassland use intensity based on multi-temporal remote sensing data and ecological indicators, Remote Sensing of Environment 198 (2017) 126--139.
\newblock \href {https://doi.org/https://doi.org/10.1016/j.rse.2017.06.003} {\path{doi:https://doi.org/10.1016/j.rse.2017.06.003}}.

\bibitem{estel2018}
S.~Estel, S.~Mader, C.~Levers, P.~Verburg, M.~Baumann, T.~Kuemmerle, Combining satellite data and agricultural statistics to map grassland management intensity in europe, Environmental Research Letters 13~(7) (2018) 074020.
\newblock \href {https://doi.org/https://doi.org/10.1088/1748-9326/aacc7a} {\path{doi:https://doi.org/10.1088/1748-9326/aacc7a}}.

\bibitem{kolecka2018}
N.~Kolecka, C.~Ginzler, R.~Pazur, B.~Price, P.~H. Verburg, Regional scale mapping of grassland mowing frequency with sentinel-2 time series, Remote Sensing 10 (2018) 1221.
\newblock \href {https://doi.org/https://doi.org/10.3390/rs10081221} {\path{doi:https://doi.org/10.3390/rs10081221}}.

\bibitem{Griffiths2019}
P.~Griffiths, C.~Nendel, J.~Pickert, P.~Hostert, Towards national-scale characterization of grassland use intensity from integrated sentinel-2 and landsat time series, Remote Sensing of Environment 238 (2020) 111124.
\newblock \href {https://doi.org/https://doi.org/10.1016/j.rse.2019.03.017} {\path{doi:https://doi.org/10.1016/j.rse.2019.03.017}}.

\bibitem{Stumpf2020}
F.~Stumpf, M.~K. Schneider, A.~Keller, A.~Mayr, T.~Rentschler, R.~G. Meuli, M.~Schaepman, F.~Liebisch, Spatial monitoring of grassland management using multi-temporal satellite imagery, Ecological Indicators 113 (2020) 106201.
\newblock \href {https://doi.org/https://doi.org/10.1016/j.ecolind.2020.106201} {\path{doi:https://doi.org/10.1016/j.ecolind.2020.106201}}.

\bibitem{sudmanns2019}
M.~Sudmanns, D.~Tiede, H.~Augustin, S.~Lang, Assessing global sentinel-2 coverage dynamics and data availability for operational earth observation (eo) applications using the eo-compass, International Journal of Digital Earth 13~(7) (2020) 768--784.
\newblock \href {https://doi.org/https://doi.org/10.1080/17538947.2019.1572799} {\path{doi:https://doi.org/10.1080/17538947.2019.1572799}}.

\bibitem{li2022}
J.~Li, C.~Li, W.~Xu, H.~Feng, F.~Zhao, H.~Long, Y.~Meng, W.~Chen, H.~Yang, G.~Yang, Fusion of optical and sar images based on deep learning to reconstruct vegetation ndvi time series in cloud-prone regions, International Journal of Applied Earth Observation and Geoinformation 112 (2022) 102818.
\newblock \href {https://doi.org/https://doi.org/10.1016/j.jag.2022.102818} {\path{doi:https://doi.org/10.1016/j.jag.2022.102818}}.

\bibitem{zeng2013}
C.~Zeng, H.~Shen, L.~Zhang, Recovering missing pixels for landsat etm+ slc-off imagery using multi-temporal regression analysis and a regularization method, Remote Sensing of Environment 131 (2013) 182--194.
\newblock \href {https://doi.org/https://doi.org/10.1016/j.rse.2012.12.012} {\path{doi:https://doi.org/10.1016/j.rse.2012.12.012}}.

\bibitem{fauvel2019}
M.~Fauvel, M.~Lopes, T.~Dubo, J.~Rivers-Moore, P.-L. Frison, N.~Gross, A.~Ouin, Prediction of plant diversity in grasslands using sentinel-1 and -2 satellite image time series, Remote Sensing of Environment 237 (2020) 111536.
\newblock \href {https://doi.org/https://doi.org/10.1016/j.rse.2019.111536} {\path{doi:https://doi.org/10.1016/j.rse.2019.111536}}.

\bibitem{sitok2021}
V.~Sitokonstantinou, A.~Koukos, T.~Drivas, C.~Kontoes, I.~Papoutsis, V.~Karathanassi, A scalable machine learning pipeline for paddy rice classification using multi-temporal sentinel data, Remote Sensing 13~(9) (2021) 1769.
\newblock \href {https://doi.org/https://doi.org/10.3390/rs13091769} {\path{doi:https://doi.org/10.3390/rs13091769}}.

\bibitem{sitokonstantinou2020sentinel}
V.~Sitokonstantinou, A.~Koutroumpas, T.~Drivas, A.~Koukos, V.~Karathanassi, H.~Kontoes, I.~Papoutsis, A sentinel based agriculture monitoring scheme for the control of the cap and food security, in: Eighth International Conference on Remote Sensing and Geoinformation of the Environment (RSCy2020), Vol. 11524, SPIE, 2020, pp. 48--59.

\bibitem{sitokonstantinou2023fuzzy}
V.~Sitokonstantinou, A.~Koukos, I.~Tsoumas, N.~S. Bartsotas, C.~Kontoes, V.~Karathanassi, Fuzzy clustering for the within-season estimation of cotton phenology, Plos one 18~(3) (2023) e0282364.

\bibitem{keay2023automated}
L.~Keay, C.~Mulverhill, N.~C. Coops, G.~McCartney, Automated forest harvest detection with a normalized planetscope imagery time series, Canadian Journal of Remote Sensing 49~(1) (2023) 2154598.

\bibitem{moreno2020}
Álvaro Moreno-Martínez, E.~Izquierdo-Verdiguier, M.~P. Maneta, G.~Camps-Valls, N.~Robinson, J.~Muñoz-Marí, F.~Sedano, N.~Clinton, S.~W. Running, Multispectral high resolution sensor fusion for smoothing and gap-filling in the cloud, Remote Sensing of Environment 247 (2020) 111901.
\newblock \href {https://doi.org/https://doi.org/10.1016/j.rse.2020.111901} {\path{doi:https://doi.org/10.1016/j.rse.2020.111901}}.

\bibitem{zhu2016}
X.~Zhu, E.~H. Helmer, F.~Gao, D.~Liu, J.~Chen, M.~A. Lefsky, A flexible spatiotemporal method for fusing satellite images with different resolutions, Remote Sensing of Environment 172 (2016) 165--177.
\newblock \href {https://doi.org/https://doi.org/10.1016/j.rse.2015.11.016} {\path{doi:https://doi.org/10.1016/j.rse.2015.11.016}}.

\bibitem{sadeh2021}
Y.~Sadeh, X.~Zhu, D.~Dunkerley, J.~P. Walker, Y.~Zhang, O.~Rozenstein, V.~Manivasagam, K.~Chenu, Fusion of sentinel-2 and planetscope time-series data into daily 3 m surface reflectance and wheat lai monitoring, International Journal of Applied Earth Observation and Geoinformation 96 (2021) 102260.
\newblock \href {https://doi.org/https://doi.org/10.1016/j.jag.2020.102260} {\path{doi:https://doi.org/10.1016/j.jag.2020.102260}}.

\bibitem{claverie2018}
M.~Claverie, J.~Ju, J.~G. Masek, J.~L. Dungan, E.~F. Vermote, J.-C. Roger, S.~V. Skakun, C.~Justice, The harmonized landsat and sentinel-2 surface reflectance data set, Remote Sensing of Environment 219 (2018) 145--161.
\newblock \href {https://doi.org/https://doi.org/10.1016/j.rse.2018.09.002} {\path{doi:https://doi.org/10.1016/j.rse.2018.09.002}}.

\bibitem{liu2019research}
C.-a. Liu, Z.-X. Chen, S.~Yun, J.-s. Chen, T.~Hasi, H.-z. PAN, Research advances of sar remote sensing for agriculture applications: A review, Journal of Integrative Agriculture 18~(3) (2019) 506--525.
\newblock \href {https://doi.org/https://doi.org/10.1016/S2095-3119(18)62016-7} {\path{doi:https://doi.org/10.1016/S2095-3119(18)62016-7}}.

\bibitem{mandal2020dual}
D.~Mandal, V.~Kumar, D.~Ratha, S.~Dey, A.~Bhattacharya, J.~M. Lopez-Sanchez, H.~McNairn, Y.~S. Rao, Dual polarimetric radar vegetation index for crop growth monitoring using sentinel-1 sar data, Remote Sensing of Environment 247 (2020) 111954.
\newblock \href {https://doi.org/https://doi.org/10.1016/j.rse.2020.111954} {\path{doi:https://doi.org/10.1016/j.rse.2020.111954}}.

\bibitem{ioannidou2022}
M.~Ioannidou, A.~Koukos, V.~Sitokonstantinou, I.~Papoutsis, C.~Kontoes, Assessing the added value of sentinel-1 polsar data for crop classification, Remote Sensing 14~(22) (2022) 5739.
\newblock \href {https://doi.org/https://doi.org/10.3390/rs14225739} {\path{doi:https://doi.org/10.3390/rs14225739}}.

\bibitem{voormansik2016}
K.~Voormansik, T.~Jagdhuber, K.~Zalite, M.~Noorma, I.~Hajnsek, Observations of cutting practices in agricultural grasslands using polarimetric sar, IEEE Journal of Selected Topics in Applied Earth Observations and Remote Sensing 9~(4) (2016) 1382--1396.
\newblock \href {https://doi.org/https://doi.org/10.1109/JSTARS.2015.2503773} {\path{doi:https://doi.org/10.1109/JSTARS.2015.2503773}}.

\bibitem{garioud2021}
A.~Garioud, S.~Valero, S.~Giordano, C.~Mallet, Recurrent-based regression of sentinel time series for continuous vegetation monitoring, Remote Sensing of Environment 263 (2021) 112419.
\newblock \href {https://doi.org/https://doi.org/10.1016/j.rse.2021.112419} {\path{doi:https://doi.org/10.1016/j.rse.2021.112419}}.

\bibitem{zhao2020}
W.~Zhao, Y.~Qu, J.~Chen, Z.~Yuan, Deeply synergistic optical and sar time series for crop dynamic monitoring, Remote Sensing of Environment 247 (2020) 111952.
\newblock \href {https://doi.org/https://doi.org/10.1016/j.rse.2020.111952} {\path{doi:https://doi.org/10.1016/j.rse.2020.111952}}.

\bibitem{shen2015missing}
H.~Shen, X.~Li, Q.~Cheng, C.~Zeng, G.~Yang, H.~Li, L.~Zhang, Missing information reconstruction of remote sensing data: A technical review, IEEE Geoscience and Remote Sensing Magazine 3~(3) (2015) 61--85.
\newblock \href {https://doi.org/https://doi.org/10.1109/MGRS.2015.2441912} {\path{doi:https://doi.org/10.1109/MGRS.2015.2441912}}.

\bibitem{li2021high}
S.~Li, L.~Xu, Y.~Jing, H.~Yin, X.~Li, X.~Guan, High-quality vegetation index product generation: A review of ndvi time series reconstruction techniques, International Journal of Applied Earth Observation and Geoinformation 105 (2021) 102640.
\newblock \href {https://doi.org/https://doi.org/10.1016/j.jag.2021.102640} {\path{doi:https://doi.org/10.1016/j.jag.2021.102640}}.

\bibitem{Yang2022}
X.~Yang, J.~Chen, Q.~Guan, H.~Gao, W.~Xia, Enhanced spatial–temporal savitzky–golay method for reconstructing high-quality ndvi time series: Reduced sensitivity to quality flags and improved computational efficiency, IEEE Transactions on Geoscience and Remote Sensing 60 (2022) 1--17.
\newblock \href {https://doi.org/https://doi.org/10.1109/TGRS.2022.3190475} {\path{doi:https://doi.org/10.1109/TGRS.2022.3190475}}.

\bibitem{roerink2000}
M.~M. G.~J.~Roerink, W.~Verhoef, Reconstructing cloudfree ndvi composites using fourier analysis of time series, International Journal of Remote Sensing 21~(9) (2000) 1911--1917.
\newblock \href {https://doi.org/https://doi.org/10.1080/014311600209814} {\path{doi:https://doi.org/10.1080/014311600209814}}.

\bibitem{chu2021}
D.~Chu, H.~Shen, X.~Guan, J.~M. Chen, X.~Li, J.~Li, L.~Zhang, Long time-series ndvi reconstruction in cloud-prone regions via spatio-temporal tensor completion, Remote Sensing of Environment 264 (2021) 112632.
\newblock \href {https://doi.org/https://doi.org/10.1016/j.rse.2021.112632} {\path{doi:https://doi.org/10.1016/j.rse.2021.112632}}.

\bibitem{zhao2023}
Y.~Zhao, P.~Hou, J.~Jiang, J.~Zhao, Y.~Chen, J.~Zhai, High-spatial-resolution ndvi reconstruction with ga-ann, Sensors 23~(4) (2023).
\newblock \href {https://doi.org/https://doi.org/10.3390/s23042040} {\path{doi:https://doi.org/10.3390/s23042040}}.

\bibitem{julien2019optimizing}
Y.~Julien, J.~A. Sobrino, Optimizing and comparing gap-filling techniques using simulated ndvi time series from remotely sensed global data, International Journal of Applied Earth Observation and Geoinformation 76 (2019) 93--111.
\newblock \href {https://doi.org/https://doi.org/10.1016/j.jag.2018.11.008} {\path{doi:https://doi.org/10.1016/j.jag.2018.11.008}}.

\bibitem{wang2019}
J.~Wang, X.~Xiao, R.~Bajgain, P.~Starks, J.~Steiner, R.~B. Doughty, Q.~Chang, Estimating leaf area index and above ground biomass of grazing pastures using sentinel-1, sentinel-2 and landsat images, ISPRS Journal of Photogrammetry and Remote Sensing 154 (2019) 189--201.
\newblock \href {https://doi.org/https://doi.org/10.1016/j.isprsjprs.2019.06.007} {\path{doi:https://doi.org/10.1016/j.isprsjprs.2019.06.007}}.

\bibitem{veloso2017}
A.~Veloso, S.~Mermoz, A.~Bouvet, T.~Le~Toan, M.~Planells, J.-F. Dejoux, E.~Ceschia, Understanding the temporal behavior of crops using sentinel-1 and sentinel-2-like data for agricultural applications, Remote Sensing of Environment 199 (2017) 415--426.
\newblock \href {https://doi.org/https://doi.org/10.1016/j.rse.2017.07.015} {\path{doi:https://doi.org/10.1016/j.rse.2017.07.015}}.

\bibitem{wang2017}
Q.~Wang, G.~A. Blackburn, A.~O. Onojeghuo, J.~Dash, L.~Zhou, Y.~Zhang, P.~M. Atkinson, Fusion of landsat 8 oli and sentinel-2 msi data, IEEE Transactions on Geoscience and Remote Sensing 55~(7) (2017) 3885--3899.
\newblock \href {https://doi.org/https://doi.org/10.1109/TGRS.2017.2683444} {\path{doi:https://doi.org/10.1109/TGRS.2017.2683444}}.

\bibitem{scarpa2018}
G.~Scarpa, M.~Gargiulo, A.~Mazza, R.~Gaetano, A cnn-based fusion method for feature extraction from sentinel data, Remote Sensing 10~(2) (2018) 236.
\newblock \href {https://doi.org/https://doi.org/10.3390/rs10020236} {\path{doi:https://doi.org/10.3390/rs10020236}}.

\bibitem{mohite2020}
J.~Mohite, S.~Sawant, A.~Pandit, S.~Pappula, Investigating the performance of random forest and support vector regression for estimation of cloud-free ndvi using sentinel-1 sar data, The International Archives of Photogrammetry, Remote Sensing and Spatial Information Sciences 43 (2020) 1379--1383.
\newblock \href {https://doi.org/https://doi.org/10.5194/isprs-archives-XLIII-B3-2020-1379-2020} {\path{doi:https://doi.org/10.5194/isprs-archives-XLIII-B3-2020-1379-2020}}.

\bibitem{pinto2022}
E.~P. Dos~Santos, D.~D. da~Silva, C.~H. do~Amaral, E.~I. Fernandes-Filho, R.~L.~S. Dias, A machine learning approach to reconstruct cloudy affected vegetation indices imagery via data fusion from sentinel-1 and landsat 8, Computers and Electronics in Agriculture 194 (2022) 106753.
\newblock \href {https://doi.org/https://doi.org/10.1016/j.compag.2022.106753} {\path{doi:https://doi.org/10.1016/j.compag.2022.106753}}.

\bibitem{pipia2019}
L.~Pipia, J.~Muñoz-Marí, E.~Amin, S.~Belda, G.~Camps-Valls, J.~Verrelst, Fusing optical and sar time series for lai gap filling with multioutput gaussian processes, Remote Sensing of Environment 235 (2019) 111452.
\newblock \href {https://doi.org/https://doi.org/10.1016/j.rse.2019.111452} {\path{doi:https://doi.org/10.1016/j.rse.2019.111452}}.

\bibitem{reyes2019}
M.~Fuentes~Reyes, S.~Auer, N.~Merkle, C.~Henry, M.~Schmitt, Sar-to-optical image translation based on conditional generative adversarial networks—optimization, opportunities and limits, Remote Sensing 11~(17) (2019) 2067.
\newblock \href {https://doi.org/https://doi.org/10.3390/rs11172067} {\path{doi:https://doi.org/10.3390/rs11172067}}.

\bibitem{meraner2020}
A.~Meraner, P.~Ebel, X.~X. Zhu, M.~Schmitt, Cloud removal in sentinel-2 imagery using a deep residual neural network and sar-optical data fusion, ISPRS Journal of Photogrammetry and Remote Sensing 166 (2020) 333--346.
\newblock \href {https://doi.org/https://doi.org/10.1016/j.isprsjprs.2020.05.013} {\path{doi:https://doi.org/10.1016/j.isprsjprs.2020.05.013}}.

\bibitem{rossberg2023globally}
T.~Ro{\ss}berg, M.~Schmitt, A globally applicable method for ndvi estimation from sentinel-1 sar backscatter using a deep neural network and the sen12tp dataset, PFG--Journal of Photogrammetry, Remote Sensing and Geoinformation Science (2023) 1--18\href {https://doi.org/https://doi.org/10.1007/s41064-023-00238-y} {\path{doi:https://doi.org/10.1007/s41064-023-00238-y}}.

\bibitem{ienco2019}
D.~Ienco, R.~Interdonato, R.~Gaetano, D.~H.~T. Minh, Combining sentinel-1 and sentinel-2 satellite image time series for land cover mapping via a multi-source deep learning architecture, ISPRS Journal of Photogrammetry and Remote Sensing 158 (2019) 11--22.
\newblock \href {https://doi.org/https://doi.org/10.1016/j.isprsjprs.2019.09.016} {\path{doi:https://doi.org/10.1016/j.isprsjprs.2019.09.016}}.

\bibitem{Ebel_and_Garnot_2023_CVPR}
P.~Ebel, V.~S.~F. Garnot, M.~Schmitt, J.~D. Wegner, X.~X. Zhu, Uncrtaints: Uncertainty quantification for cloud removal in optical satellite time series, in: Proceedings of the IEEE/CVF Conference on Computer Vision and Pattern Recognition (CVPR) Workshops, 2023, pp. 2086--2096.

\bibitem{rusvurm2018}
M.~Ru{\ss}wurm, M.~K{\"o}rner, Multi-temporal land cover classification with sequential recurrent encoders, ISPRS International Journal of Geo-Information 7~(4) (2018) 129.
\newblock \href {https://doi.org/https://doi.org/10.3390/ijgi7040129} {\path{doi:https://doi.org/10.3390/ijgi7040129}}.

\bibitem{rovberg_and_schmitt_2024}
T.~Roßberg, M.~Schmitt, Dense ndvi time series by fusion of optical and sar-derived data, IEEE Journal of Selected Topics in Applied Earth Observations and Remote Sensing 17 (2024) 7748--7758.
\newblock \href {https://doi.org/https://doi.org/10.1109/JSTARS.2024.3379838} {\path{doi:https://doi.org/10.1109/JSTARS.2024.3379838}}.

\bibitem{ali2017}
I.~Ali, B.~Barrett, F.~Cawkwell, S.~Green, E.~Dwyer, M.~Neumann, Application of repeat-pass terrasar-x staring spotlight interferometric coherence to monitor pasture biophysical parameters: Limitations and sensitivity analysis, IEEE Journal of Selected Topics in Applied Earth Observations and Remote Sensing 10~(7) (2017) 3225--3231.
\newblock \href {https://doi.org/https://doi.org/10.1109/JSTARS.2017.2679761} {\path{doi:https://doi.org/10.1109/JSTARS.2017.2679761}}.

\bibitem{devroey2021}
M.~De~Vroey, J.~Radoux, P.~Defourny, Grassland mowing detection using sentinel-1 time series: Potential and limitations, Remote Sensing 13~(3) (2021) 348.
\newblock \href {https://doi.org/https://doi.org/10.3390/rs13030348} {\path{doi:https://doi.org/10.3390/rs13030348}}.

\bibitem{tamm2016}
T.~Tamm, K.~Zalite, K.~Voormansik, L.~Talgre, Relating sentinel-1 interferometric coherence to mowing events on grasslands, Remote Sensing 8 (2016) 802.
\newblock \href {https://doi.org/https://doi.org/10.3390/rs8100802} {\path{doi:https://doi.org/10.3390/rs8100802}}.

\bibitem{reinermann2022detection}
S.~Reinermann, U.~Gessner, S.~Asam, T.~Ullmann, A.~Schucknecht, C.~Kuenzer, Detection of grassland mowing events for germany by combining sentinel-1 and sentinel-2 time series, Remote Sensing 14~(7) (2022) 1647.
\newblock \href {https://doi.org/https://doi.org/10.3390/rs14071647} {\path{doi:https://doi.org/10.3390/rs14071647}}.

\bibitem{devroey2022}
M.~{De Vroey}, L.~{de Vendictis}, M.~Zavagli, S.~Bontemps, D.~Heymans, J.~Radoux, B.~Koetz, P.~Defourny, Mowing detection using sentinel-1 and sentinel-2 time series for large scale grassland monitoring, Remote Sensing of Environment 280 (2022) 113145.
\newblock \href {https://doi.org/https://doi.org/10.1016/j.rse.2022.113145} {\path{doi:https://doi.org/10.1016/j.rse.2022.113145}}.

\bibitem{schwieder2021}
M.~Schwieder, M.~Wesemeyer, D.~Frantz, K.~Pfoch, S.~Erasmi, J.~Pickert, C.~Nendel, P.~Hostert, Mapping grassland mowing events across germany based on combined sentinel-2 and landsat 8 time series, Remote Sensing of Environment 269 (2022) 112795.
\newblock \href {https://doi.org/https://doi.org/10.1016/j.rse.2021.112795} {\path{doi:https://doi.org/10.1016/j.rse.2021.112795}}.

\bibitem{taravat2019}
A.~Taravat, M.~Wagner, N.~Oppelt, Automatic grassland cutting status detection in the context of spatiotemporal sentinel-1 imagery analysis and artificial neural networks, Remote Sensing 11~(6) (2019) 711.
\newblock \href {https://doi.org/https://doi.org/10.3390/rs11060711} {\path{doi:https://doi.org/10.3390/rs11060711}}.

\bibitem{komisarenko2022}
V.~Komisarenko, K.~Voormansik, R.~El~Shawi, S.~Sakr, Exploiting time series of sentinel-1 and sentinel-2 to detect grassland mowing events using deep learning with reject region, Scientific Reports 12 (2022) 983.
\newblock \href {https://doi.org/https://doi.org/10.1038/s41598-022-04932-6} {\path{doi:https://doi.org/10.1038/s41598-022-04932-6}}.

\bibitem{lobert2021}
F.~Lobert, A.-K. Holtgrave, M.~Schwieder, M.~Pause, J.~Vogt, A.~Gocht, S.~Erasmi, Mowing event detection in permanent grasslands: Systematic evaluation of input features from sentinel-1, sentinel-2, and landsat 8 time series, Remote Sensing of Environment 267 (2021) 112751.
\newblock \href {https://doi.org/https://doi.org/10.1016/j.rse.2021.112751} {\path{doi:https://doi.org/10.1016/j.rse.2021.112751}}.

\bibitem{Lange2022}
M.~Lange, H.~Feilhauer, I.~K{\"u}hn, D.~Doktor, Mapping land-use intensity of grasslands in germany with machine learning and sentinel-2 time series, Remote Sensing of Environment 277 (2022) 112888.
\newblock \href {https://doi.org/https://doi.org/10.1016/j.rse.2022.112888} {\path{doi:https://doi.org/10.1016/j.rse.2022.112888}}.

\bibitem{Holtgrave2023}
A.-K. Holtgrave, F.~Lobert, S.~Erasmi, N.~Röder, B.~Kleinschmit, Grassland mowing event detection using combined optical, sar, and weather time series, Remote Sensing of Environment 295 (2023) 113680.
\newblock \href {https://doi.org/https://doi.org/10.1016/j.rse.2023.113680} {\path{doi:https://doi.org/10.1016/j.rse.2023.113680}}.

\bibitem{lithuania_climatic_regions}
A.~Laurinavi{\v{c}}ius, D.~{\v{C}}ygas, A.~Vaitkus, T.~Ratkevi{\v{c}}ius, Climatic regioning of lithuania from the point of view of winter road maintenance, in: XIV th International Winter Road Congress, 2014, pp. 1--9.

\bibitem{Pinheiro2022}
M.~Pinheiro, N.~Miranda, A.~Recchia, A.~Cotrufo, N.~Franceschi, R.~Piantanida, K.~Schmidt, C.~Gisinger, G.~Hajduch, P.~Vincent, Sentinel-1 instruments status and product performance update for 2022, in: EUSAR 2022; 14th European Conference on Synthetic Aperture Radar, VDE, 2022, pp. 1--5.

\bibitem{sen2cor_new}
J.~Louis, B.~Pflug, M.~Main-Knorn, V.~Debaecker, U.~Mueller-Wilm, R.~Q. Iannone, E.~Giuseppe~Cadau, V.~Boccia, F.~Gascon, Sentinel-2 global surface reflectance level-2a product generated with sen2cor, in: IGARSS 2019 - 2019 IEEE International Geoscience and Remote Sensing Symposium, 2019, pp. 8522--8525.
\newblock \href {https://doi.org/https://doi.org/10.1109/IGARSS.2019.8898540} {\path{doi:https://doi.org/10.1109/IGARSS.2019.8898540}}.

\bibitem{datacube_2022}
T.~Drivas, V.~Sitokonstantinou, I.~Tsardanidis, A.~Koukos, C.~Kontoes, V.~Karathanassi, A data cube of big satellite image time-series for agriculture monitoring, in: 2022 IEEE 14th Image, Video, and Multidimensional Signal Processing Workshop (IVMSP), 2022, pp. 1--5.
\newblock \href {https://doi.org/https://doi.org/10.1109/IVMSP54334.2022.9816291} {\path{doi:https://doi.org/10.1109/IVMSP54334.2022.9816291}}.

\bibitem{tarrio2020}
K.~Tarrio, X.~Tang, J.~G. Masek, M.~Claverie, J.~Ju, S.~Qiu, Z.~Zhu, C.~E. Woodcock, Comparison of cloud detection algorithms for sentinel-2 imagery, Science of Remote Sensing 2 (2020) 100010.
\newblock \href {https://doi.org/https://doi.org/10.1016/j.srs.2020.100010} {\path{doi:https://doi.org/10.1016/j.srs.2020.100010}}.

\bibitem{WATZIG2023113577}
C.~Watzig, A.~Schaumberger, A.~Klingler, A.~Dujakovic, C.~Atzberger, F.~Vuolo, Grassland cut detection based on sentinel-2 time series to respond to the environmental and technical challenges of the austrian fodder production for livestock feeding, Remote Sensing of Environment 292 (2023) 113577.
\newblock \href {https://doi.org/https://doi.org/10.1016/j.rse.2023.113577} {\path{doi:https://doi.org/10.1016/j.rse.2023.113577}}.

\bibitem{akima1970}
H.~Akima, A new method of interpolation and smooth curve fitting based on local procedures, Journal of the ACM (JACM) 17~(4) (1970) 589--602.
\newblock \href {https://doi.org/https://doi.org/10.1145/321607.321609} {\path{doi:https://doi.org/10.1145/321607.321609}}.

\bibitem{bontemps2022sen4cap}
S.~Bontemps, K.~Bajec, C.~Cara, P.~Defourny, L.~de~Vendictis, D.~Heymans, L.~Kucera, P.~Malcorps, G.~Milcinski, L.~Nicola, J.~Slacikova, M.~Taymans, F.~Tutunaru, C.~Udroiu, Sen4cap - sentinels for common agricultural policy., Syst. Softw. User Manual. (2022).

\bibitem{noor2014filling}
M.~Noor, A.~Yahaya, N.~A. Ramli, A.~M. Al~Bakri, Filling missing data using interpolation methods: Study on the effect of fitting distribution, Key Engineering Materials 594 (2014) 889--895.
\newblock \href {https://doi.org/https://doi.org/10.4028/www.scientific.net/kem.594-595.889.} {\path{doi:https://doi.org/10.4028/www.scientific.net/kem.594-595.889.}}

\bibitem{hardy2021sen2grass}
T.~Hardy, L.~Kooistra, M.~Domingues~Franceschini, S.~Richter, E.~Vonk, G.~van~den Eertwegh, D.~Van~Deijl, Sen2grass: A cloud-based solution to generate field-specific grassland information derived from sentinel-2 imagery, AgriEngineering 3~(1) (2021) 118--137.
\newblock \href {https://doi.org/https://doi.org/10.3390/agriengineering3010008} {\path{doi:https://doi.org/10.3390/agriengineering3010008}}.

\end{thebibliography}





\end{document}